\documentclass[times,twocolumn,final]{elsarticle}

\usepackage{medima}
\usepackage{framed,multirow}

\usepackage{amssymb}
\usepackage{latexsym}

\usepackage{url}
\usepackage{xcolor}

\usepackage{hyperref}

\definecolor{newcolor}{rgb}{.8,.349,.1}

\journal{Medical Image Analysis}

\usepackage{graphicx}
\usepackage{amsmath}
\usepackage{booktabs}
\usepackage{blindtext}
\usepackage{tabularx}
\usepackage{caption, subcaption, sidecap}
\usepackage[export]{adjustbox}

\usepackage[shortlabels]{enumitem}

\usepackage{float}

\usepackage{graphicx}
\usepackage{subcaption}
\usepackage{tikz}
\usepackage{multirow}
\usepackage{cleveref}

\begin{document}

\verso{Given-name Surname \textit{et~al.}}

\begin{frontmatter}

\title{SAR-RARP50: Segmentation of surgical instrumentation and Action Recognition on Robot-Assisted Radical Prostatectomy Challenge}%


\author[1]{Dimitrios \snm{Psychogyios}}
\ead{dimitris.psychogyios.19@ucl.ac.uk}


\author[1]{Emanuele \snm{Colleoni}}
\ead{ emanuele.colleoni.19@ucl.ac.uk}


\author[1]{Beatrice~Van  \snm{Amsterdam}}
\ead{beatrice.amsterdam.18@ucl.ac.uk}


\author[3,4]{Chih-Yang  \snm{Li}}
\author[3,4]{Shu-Yu  \snm{Huang}}


\author[5]{Yuchong  \snm{Li}}
\author[6]{Fucang  \snm{Jia}}

\author[7]{Baosheng  \snm{Zou}}
\author[7]{Guotai  \snm{Wang}}


\author[8]{Yang \snm{Liu}}
\author[8]{Maxence \snm{Boels}}
\author[8]{Jiayu \snm{Huo}}
\author[8]{Rachel \snm{Sparks}}
\author[8]{Prokar \snm{Dasgupta}}
\author[8]{Alejandro \snm{Granados}}
\author[8]{Sébastien \snm{Ourselin}}

\author[9]{Mengya \snm{Xu}}
\author[9]{An \snm{Wang}}
\author[9]{Yanan \snm{Wu}}
\author[9]{Long \snm{Bai}} 
\author[9]{Hongliang \snm{Ren}}

\author[10]{Atsushi \snm{Yamada}}
\author[10]{Yuriko \snm{Harai}}
\author[10]{Yuto \snm{Ishikawa}}
\author[10]{Kazuyuki \snm{Hayashi}}

\author[11]{Jente \snm{Simoens}}
\author[11]{Pieter De\snm{Backer}}
\author[12]{Francesco \snm{Cisternino}}
\author[12]{Gabriele \snm{Furnari}}
\author[11]{Alex \snm{Mottrie}}
\author[12]{Federica \snm{Ferraguti}}

\author[13]{Satoshi \snm{Kondo}}
\author[14]{Satoshi \snm{Kasai}}
\author[15]{Kousuke \snm{Hirasawa}}

\author[16,17]{Soohee \snm{Kim}}
\author[16]{Seung Hyun \snm{Lee}}
\author[16,18]{Kyu Eun \snm{Lee}}
\author[16,18]{Hyoun-Joong \snm{Kong}}

\author[19]{Kui \snm{Fu}}
\author[19]{Chao \snm{Li}}
\author[19]{Shan \snm{An}}

\author[20]{Stefanie \snm{Krell}}
\author[20]{Sebastian \snm{Bodenstedt}}

\author[21]{Nicolas \snm{Ayobi}}
\author[21]{Alejandra \snm{Perez}}
\author[21]{Santiago \snm{Rodriguez}}
\author[21]{Juanita \snm{Puentes}}
\author[21]{Pablo \snm{Arbelaez}}

\author[22]{Omid \snm{Mohareri}}

\author[1]{Danail  \snm{Stoyanov}}



\address[1]{Univercity College London, London, United Kingdom}

\address[3]{Taiwan AI Academy, New Taipei City, Taiwan}
\address[4]{National Taiwan University, Taipei, Taiwan}

\address[5]{Shenzhen Institute of Advanced Technology, Chinese Academy of Sciences, Shenzhen, China}
\address[6]{University of Chinese Academy of Sciences, Beijing, China}

\address[7]{University of Electronic Science and Technology of China, Sichuan, China}

\address[8]{King’s College London, London, United Kingdom}
\address[9]{The Chinese University of Hong Kong, Hong Kong, China}
\address[10]{National Cancer Center Hospital East, Chiba, Japan}
\address[11]{ORSI Academy, Melle, Belgium}
\address[12]{University of Modena and Reggio Emilia, Modena, Italy}

\address[13]{Muroran Institute of Technology, Hokkaido, Japan}
\address[14]{Niigata University of Health and Welfare, Niigata, Japan}
\address[15]{Konica Minolta, Inc, Osaka, Japan}


\address[16]{Seoul National University Hospital, Seoul, Korea}
\address[17]{Seoul National University, Seoul, Korea}
\address[18]{Seoul National University College of Medicine, Seoul, Korea}


\address[19]{JD Health International Inc., Beijing, China}


\address[20]{National Center for Tumor Diseases (NCT), Dresden, Germany}

\address[21]{University of Los Andes, Bogota, Colombia}

\address[22]{Intuitive Surgical, Inc., Sunnyvale, California, United States}

\received{1 May 2013}
\finalform{10 May 2013}
\accepted{13 May 2013}
\availableonline{15 May 2013}
\communicated{S. Sarkar}

\begin{abstract}

Surgical tool segmentation and action recognition are fundamental building blocks in many computer-assisted intervention applications, ranging from surgical skills assessment to decision support systems. Nowadays, learning-based action recognition and segmentation approaches outperform classical methods, relying, however, on large, annotated datasets. Furthermore, action recognition and tool segmentation algorithms are often trained and make predictions in isolation from each other, without exploiting potential cross-task relationships. With the EndoVis 2022 SAR-RARP50 challenge, we release the first multimodal, publicly available, in-vivo, dataset for surgical action recognition and semantic instrumentation segmentation, containing 50 suturing video segments of Robotic Assisted Radical Prostatectomy (RARP). The aim of the challenge is twofold. First, to enable researchers to leverage the scale of the provided dataset and develop robust and highly accurate single-task action recognition and tool segmentation approaches in the surgical domain. Second, to further explore the potential of multitask-based learning approaches and determine their comparative advantage against their single-task counterparts. A total of 12 teams participated in the challenge, contributing 7 action recognition methods, 9 instrument segmentation techniques, and 4 multitask approaches that integrated both action recognition and instrument segmentation. The complete SAR-RARP50 dataset is available at \url{https://rdr.ucl.ac.uk/projects/SAR-RARP50_Segmentation_of_surgical_instrumentation_and_Action_Recognition_on_Robot-Assisted_Radical_Prostatectomy_Challenge/191091}

\end{abstract}


\end{frontmatter}

\section{Introduction}

Understanding surgical processes and the surgical environment, e.g. the location of anatomy and instrumentation is essential for developing modern clinical support systems \cite{chadebecq2020computer}. As an example, the analysis of surgical motion at a fine-grained scale finds application in multiple contexts such as surgical skill assessment \cite{gao2014jhu} and automation of surgical motion \cite{nagy2019dvrk}.

Studies evaluating action recognition systems have primarily relied on small and constrained datasets of surgical training sessions \cite{gao2014jhu}, which fail to capture the diversity and complexity of real-world surgical scenarios. Similarly, while advancements in surgical instrumentation segmentation methods for applications such as surgical navigation \cite{islam2019learning} and visualization systems \cite{wang2022neural} are underway, their development and evaluation often rely on datasets acquired under controlled conditions \cite{allan20192017,allan20202018}. These conditions may differ significantly from the dynamic and unpredictable nature of real surgical video scenarios. Because of the domain gap between training data and real surgical video, learning-based approaches have been known to underperform when processing real surgical video. 


These limitations primarily arise from the challenges associated with collecting real surgical data at scale. Ethical, regulatory, and legal constraints, as well as the logistical hurdles of managing and coordinating multi-centre datasets, make it difficult to collect and annotate large amounts of surgical data. Additionally, the non-trivial standardization of annotation criteria for different procedures and the time-consuming, expensive, and error-prone nature of manual labelling further hinder the production of new data.

To mitigate data scarcity and facilitate advancements in surgical vision, this challenge provides labelled datasets for training and validating deep-learning models in real-world surgeries. The provided videos cover intricate anatomies, dynamic camera movements, diverse and challenging lighting conditions, and the presence of blood and occlusions. Furthermore, the variability in both action sequence and execution strategy across videos enables assessment of both action recognition and surgical instrumentation segmentation in real-world scenarios. As such, the dataset's multi-modal nature allows participants to exploit intrinsic relations between action recognition and instrument segmentation, potentially improving predictions for either task.


\section{Related work}

\subsection{Surgical action recognition} \label{arec}
Automatic recognition of surgical gestures is difficult due to the complexity and variability of surgical activities. Data variability is not only explained by user-specific operative style and skill level but it is also linked to environmental conditions such as the type of intervention and the patient-specific anatomy \cite{dergachyova2017knowledge}. State-of-the-art methods tackle this complex problem using deep neural networks to leverage high computational power and substantial amounts of training data. Several studies were focused on robust modelling of temporal information through hierarchical temporal \cite{lea2016temporal} or graph \cite{kadkhodamohammadi2022patg} convolutions, recurrent modules \cite{jin2017sv} or attention mechanisms \cite{gao2021trans,nwoye2022rendezvous}. Network understanding of surgical processes can be strengthened with the integration of different sensor data carrying complementary information \cite{long2021relational,van2022gesture}. Additionally, multi-task learning using appropriate auxiliary tasks, such as estimation of surgical tool trajectory \cite{qin2020davincinet} or surgical skill assessment \cite{wang2021towards}, has also shown a potential to improve recognition performance.
Motivated by these results, the proposed challenge aims to investigate the effectiveness of multi-task learning in complex real-case scenarios.

\subsection{Surgical instrumentation Segmentation} 
Accurately segmenting surgical instruments is crucial for comprehending the surgical scene through video analysis. For several years, Fully Convolutional Neural Networks (FCNN) dominated the field, mainly leveraging U-net-based architectures \cite{ronneberger2015u} and incorporating pre-trained ResNet backbones. Modifications to the decoder allowed for parallel binary and semantic segmentation \cite{allan20192017}, integration of localization and classification heads in a multi-task fashion \cite{fathabadi2021multi, ciaparrone2020comparative}, or holistic nesting of features extracted at different scales in the encoder \cite{garcia2017toolnet}. Other recent approaches also tested positive learning pipelines, where image labels different from segmentation are used to improve performance in an unsupervised fashion~ \cite{psychogyios2022msdesis}. Recently, transformers~ \cite{vaswani2017attention} showed outperforming results compared to FCNNs~ \cite{zhao2022trasetr, shamshad2023transformers}, thus establishing a new baseline for surgical tool and instrument segmentation. Adding temporal information also showed to improve model performance over standard pipelines~ \cite{jin2019incorporating, kanakatte2020surgical} thanks to their ability to refine predictions based on past knowledge. Generative and adversarially trained models have also been studied both as a tool for data generation and augmentation~ \cite{colleoni2022ssis} as well as to refine segmentation prediction with a discriminative loss~ \cite{kalia2021co, sahu2021simulation}.

\section{Challenge Description}
\subsection{Tasks}

\subsubsection{Task 1: Action recognition}
The first sub-task consists of decomposing real surgical demonstrations into fine-grained temporal segments and classifying them into a pre-defined set of action classes. 
State-of-the-art approaches perform well in controlled environments, with limited noise and action sequence variability \cite{gao2014jhu, stein2013combining}, so the goal is to find an accurate solution for complex videos of real surgical interventions. 

\subsubsection{Task 2: Surgical instrumentation semantic segmentation}
The second sub-task involves processing red-green-blue (RGB) images and assigning semantic labels at the pixel level. This process results in image masks of prominent objects such as surgical tool parts but also thin and small tools such as surgical clips and suturing threads and needles. Currently, machine learning approaches are optimized to segment tool parts and achieve great accuracy in datasets captured under controlled conditions, such as ex-vivo or porcine environments \cite{colleoni2020synthetic, allan20192017, allan20202018}. The goal of this task is therefore to investigate how such models perform when applied to data with challenging lighting conditions, camera focus, and blood occlusions.

\subsubsection{Task 3: Multitask}

The final sub-task focuses on predicting surgical action labels and surgical instrumentation segmentation masks simultaneously, using only the surgical video as input. While single-task methods can make accurate predictions, they limit the potential for context-aware optimizations based on other modalities. Integrating multimodal information during optimization could enable better modelling of the surgical environment, leading to more robust and accurate predictions. Additionally, multi-task architectures could enable faster inference by sharing network components between the two tasks. However, multitask learning poses challenges due to differences in sampling rates for each modality and the need to balance learning objectives during optimization. The goal of this challenge is to overcome these difficulties and train multimodal approaches to effectively address both tasks. This sub-challenge allows for either a cascade of single-task networks, where the second network utilizes predictions from the first or networks that employ shared architecture components to estimate both tasks.

\subsection{Evaluation}

\subsubsection{Action recognition}\label{sec:acrec_metrics}
We assess the performance of action recognition algorithms at 10 Hz, using the Frame-wise accuracy and the segmental F1@K.

The Frame-wise accuracy \ref{loss:fwa} is calculated by dividing the number of frames that are correctly classified by the total number of frames in the video. Such a metric can be used to identify how well a model performs at classifying a specific class.

\begin{equation}
{FWA}_i = \frac{\text{\#correctly classified frames}}{\text{\# frames in the video i}}
\label{loss:fwa}
\end{equation}

To assess the model's temporal performance, we also measure the Segmental F1@k (\ref{loss:f1@10}) score which is designed to penalize out-of-order predictions and over-segmentation. This metric assesses the temporal overlap between predictions and target segments with reduced sensitivity to slight temporal shifts, compensating for annotation noise around the segment boundaries. The segmental F1@K is computed as the IoU overlap score between each predicted segment and the corresponding target segment of the same class. That prediction is considered a true positive (TP) if the IoU is above a threshold T = k/100, otherwise, it is a false positive (FP). TPs and FPs are then used to compute the final F1 score. For the SAR-RARP50 challenge, we set $K=10$.

\begin{equation}
segmentalF1@K = \frac{2 \times (\text{precision} \times \text{recall})}{(\text{precision} + \text{recall})}
\label{loss:f1@10}
\end{equation}

We compute (\ref{loss:fwa}), (\ref{loss:f1@10}) for every dataset $i$ in the test set and we average results across the test set (\ref{loss:fwa_avg}), (\ref{loss:f1@10_avg}).

\begin{equation}
FWA_{avg} = \frac{1}{M} \sum_{k=1}^{M}FWA_{i}
\label{loss:fwa_avg}
\end{equation}

\begin{equation}
F1@10_{avg} = \frac{1}{M} \sum_{k=1}^{M}F1@10_{i}
\label{loss:f1@10_avg}
\end{equation}

With $M=10$ for SAR-RARP50
The final score for the action recognition task, which is used to rank participants is defined in (\ref{loss:ar})

\begin{equation}
Score_{ar} = \sqrt{FWA_{avg} * F1@10_{avg}}
\label{loss:ar}
\end{equation}

After the challenge completion, we analyzed the ranking stability among teams. Since such analysis required us to assign a score to each video, we computed a per-video Action recognition score as described in \ref{loss:var}

\begin{equation}
AR\_Stability_i = \sqrt{FWA_{i} * F1@10_{i}}
\label{loss:var}
\end{equation}

\subsubsection{Surgical instrumentation segmentation}\label{sec:seg_metrics}
Segmentation predictions are evaluated at 1FPS, at 1920x1080 resolution using two metrics:

\textit{mean Intersection over Union}: intersection over union~(IoU) is a commonly used metric, used to evaluate segmentation methods at the pixel level. It measures the overlapping between a model prediction and the target mask. For each frame \textit{j} in video \textit{i}, IoU for semantic class \textit{k}~$\in$~\textit{K} can be computed as:

\begin{equation}
IoU_{ijk} = \frac{GT_{ijk}~\cap~Prediction_{ijk}}{GT_{ijk}~\cup~Prediction_{ijk}}
\end{equation}

where \textit{K} is the set of all the semantic classes while GT and Prediction are the target and predicted masks relative to frame \textit{j}, respectively.
The mean Intersection over Union (mIoU) for a given frame over all the semantic classes is then computed as:

\begin{equation}
mIoU_{ij} = \frac{1}{K}\sum_{k=1}^{K}IoU_{ijk}
\end{equation}

Similarly, the mIoU for a video \textit{i} can be computed as:

\begin{equation}
mIoU_{i} = \frac{1}{J}\sum_{j=1}^{J}mIoU_{ij}
\end{equation}

and, finally, the final mIoU score is computed over all the test videos as:

\begin{equation}
mIoU = \frac{1}{I}\sum_{i=1}^{I}mIoU_{i}
\end{equation}

Although being extensively used in literature, in IoU computation, all pixels are weighted equally and, as such, a misclassified pixel close to the reference mask boundary will have the same impact as a pixel erroneously predicted far from its class location. Considering this metric alone for model evaluation may lead to a superficial analysis of the models' results and should be supported by other metrics that can provide for IoU's deficiency.

\textit{mean Normalized Surface Dice}:
The Normalized Surface Dice (NSD) computes the number of predicted boundary pixels whose distance from the closest boundary pixel in the target mask is shorter than a given distance threshold. Compared to IoU, NSD provides a more weighted estimate of the prediction quality by not penalizing false positives close to the target mask, although it would not penalize little misclassifications within the prediction boundaries. For this reason, we use both IoU and NSD to evaluate segmentation models.
The NSD used for SAR-RARP50 challenge follows \cite{seidlitz2022robust} implementation and it is defined as follows:

For each frame \textit{j} in each video \textit{i}, $\beta^{Pred}_{ijk}$ is defined as the set of boundary pixels for prediction ijk, where \textit{k}~$\in$~\textit{K} is the corresponding semantic class. In the same way, $\beta^{Target}_{ijk}$ can be defined as the set of boundary pixels in the target segmentation map. From these two sets, we define two additional sets $\Delta^{Pred}_{ijk}$ and $\Delta^{Target}_{ijk}$ that, for each boundary pixel in the prediction mask, contains the nearest neighbour distance to the target mask and vice-versa. Finally, two sub-sets $\delta^{Pred}_{ijk}$ and $\delta^{Target}_{ijk}$ can be constructed by filtering out all the distances higher than a given threshold $\tau$.

The NSD for class k is defined as:

\begin{equation}
NSD_{ijk} = \frac{\| \delta^{Pred}_{ijk} \| + \| \delta^{Target}_{ijk} \|}{\| \Delta^{Target}_{ijk} \| + \|\Delta^{Target}_{ijk}  \|}
\end{equation}

From this, we can calculate the mean NSD over all classes in a given frame \textit{j} as:

\begin{equation}
mNSD_{ij} = \frac{1}{K}\sum_{k=1}^{K}NSD_{ijk}
\end{equation}

and over all frames in video \textit{i} as:

\begin{equation}
mNSD_{i} = \frac{1}{J}\sum_{j=1}^{J}mNSD_{ij}
\end{equation}

The overall mNSD score for each submission is computed as:

\begin{equation}
mNSD = \frac{1}{I}\sum_{i=1}^{I}mNSD_{i}
\end{equation}

\textit{Segmentation score}: The final score for the segmentation sub-challenge is given by the following formula:

\begin{equation}\label{eq: seg_score}
Score_{s} = \sqrt{mIoU * mNSD}
\end{equation}

Similar to the Action recognition sub-challenge, we analyzed the ranking stability among teams for the task of semantic segmentation. We compute a per video $i$ semantic segmentation score as described in \ref{loss:vss}

\begin{equation}
SS\_Stability_{i} = \sqrt{mNSD_{i} * mIoU_{i}}
\label{loss:vss}
\end{equation}

\subsubsection{Multitask}\label{sec:mul_metrics}
Similarly to equation \ref{eq: seg_score}, the final score for multitask model evaluation is defined as:

\begin{equation}\label{eq: mt_score}
Score_{mt} = \sqrt{Score_{ar} * Score_{s}}
\end{equation}

\section{Dataset}
The SAR-RARP50 video dataset includes action and surgical instrumentation labels for video segments recorded during 50 Robot-Assisted Radical Prostatectomies (RARP). 
The selected segments focus on the suturing of the dorsal vascular complex (DVC), an array of veins and arteries that is sutured to keep bleeding under control after the connection of the prostate to the bladder and urethra is cut. The data were collected at the University College Hospital at Westmoreland Street, London, UK, and included operations performed by 8 surgeons with different surgical seniority (experienced consultant, senior registrar, and junior registrar).
SAR-RARP50 is a superset of the RARP45 dataset \cite{van2022gesture} and expands it by adding 5 more operations and introducing surgical instrumentation segmentation reference masks for all videos at a rate of 1Hz.
The provided videos differ in terms of lighting conditions, colour (due to variations in the light source), length, amount of blood present in the scene, and image clarity (due to fluids ending up on the camera lens).

\subsection{Data collection and pre-processing}
All provided surgical operations were performed using a DaVinci® Si robot (Intuitive Surgical Inc, CA), and data recording started after the endoscope was placed inside the patient's abdomen. The dVLogger device was used to record each of the two stereo channels individually at 60 frames per second at 1080i resolution. Video was encoded and stored in an external device together with files containing frame timestamp information. After data acquisition, the timestamp files were used to time-synchronize the two stereo video channels and rewrite them using a common, fixed frame rate. The resulting video files were further processed to remove interlace artefacts.

\subsection{Action recognition annotation protocol}

A dictionary of seven bi-manual gestures and a background class was collaboratively designed by an expert surgeon and a machine learning researcher to facilitate manual segmentation of DVC suturing demonstrations. The gesture labels are listed in Table \ref{table:gesture_list}. The annotation process involved loading each video to custom annotation software, allowing per-frame label assignment.  Annotations were generated by a trained machine learning researcher as there was no disagreement between clinical and non-clinical interpretations of the surgical gestures used in the study. Labels were assigned to frames where a new surgical action began, effectively annotating all frames between action transitions. Annotations were provided to challenge participants as a frame list and corresponding label list at a rate of 10 Hz.

\begin{table}[]
	\centering
	\caption{RARP-50 dataset gesture list.}
	\renewcommand{\arraystretch}{1.2}
	\begin{tabular}{|c|l|l|}
	\hline
	\textbf{ID}	& \textbf{\ \ \ \ Gesture description} \\
	\hline
	G0 & \ \ Other \\
    G1 & \ \ Picking-up the needle \\
    G2 & \ \ Positioning the needle tip \\
    G3 & \ \ Pushing the needle through the tissue \\
    G4 & \ \ Pulling the needle out of the tissue \\
    G5 & \ \ Tying a knot \\
    G6 & \ \ Cutting the suture \\
    G7 & \ \ Returning/dropping the needle \\
	\hline
	\end{tabular}
	\label{table:gesture_list}
\end{table}

\begin{figure}
    \centering
    \includegraphics[width=\linewidth]{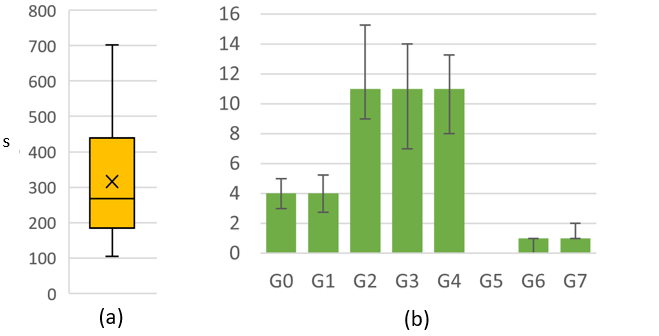}
    \caption{RARP-45 statistics. (a) Task duration variability (reported in seconds). The average duration is about 5 minutes, with large variability ranging from about 2 to 12 minutes. (b) Class distribution per sequence. Each bin represents the median class frequency over interventions, and error bars mark the 25th and 75th quantiles. Class G5 is absent in more than 75\% of the interventions.}
    \label{fig:wspcount}
\end{figure}

The dataset presents challenges due to significant variability across different operations and surgical gestures. This diversity derives from factors such as the duration of each action(Fig. \ref{fig:wspcount}a), their frequency (Fig. \ref{fig:wspcount}b) and ordering. While some variability is linked to operator-dependent factors like surgical style and robotic experience, it is also influenced by patient-specific anatomical structure and its response to manipulation. Real-case variability factors, such as unexpected or excessive bleeding, can prompt multiple gesture attempts or alter the surgical strategy. 

\subsection{Semantic segmentation annotation protocol}\label{sec:seg_protocol}

The dataset annotates nine semantic classes labelling all areas of the surgical scene that correspond to surgical instrumentation. The nine classes cluster non-tool objects into the following categories: suturing needle, suture thread, surgical clip, suction tool, needle holder, and catheter. To allow semantic propagation across different types of robotics tools, semantic classes are assigned at part level and include: shaft, wrist, and claspers.

\begin{figure*}[hbt!]
    \centering 
    \begin{subfigure}{0.33\textwidth}
      \includegraphics[width=\linewidth]{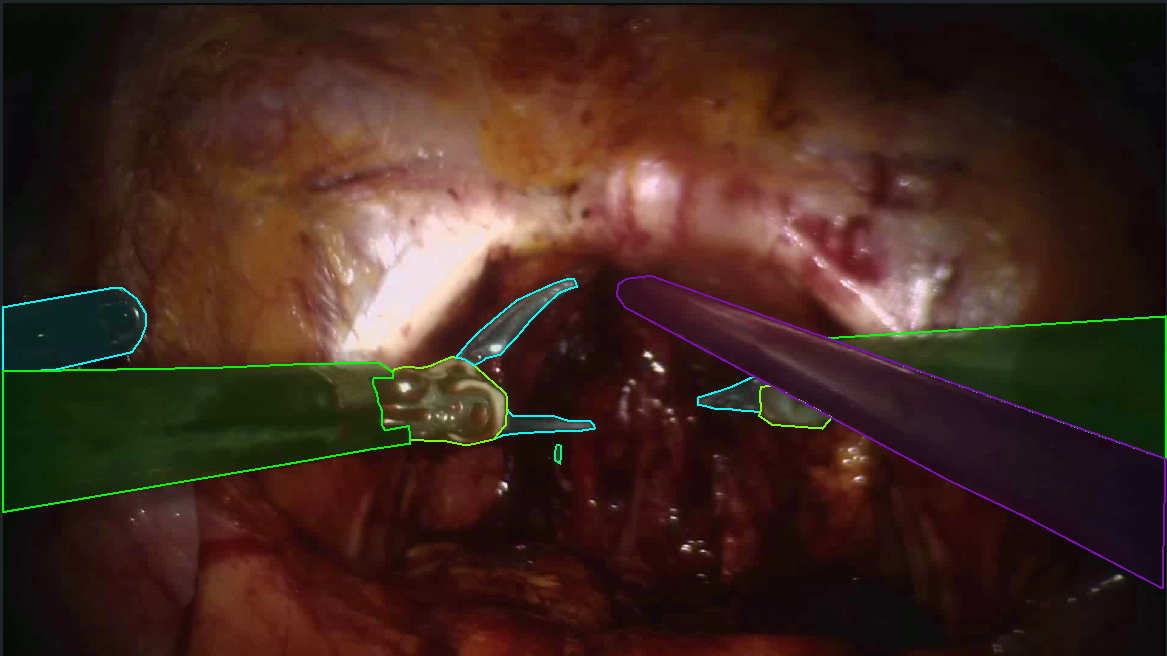}
      \caption{Labels of overlapping objects}
      \label{fig:protocol_seg_1}
    \end{subfigure}\hfil 
    \begin{subfigure}{0.33\textwidth}
      \includegraphics[width=\linewidth]{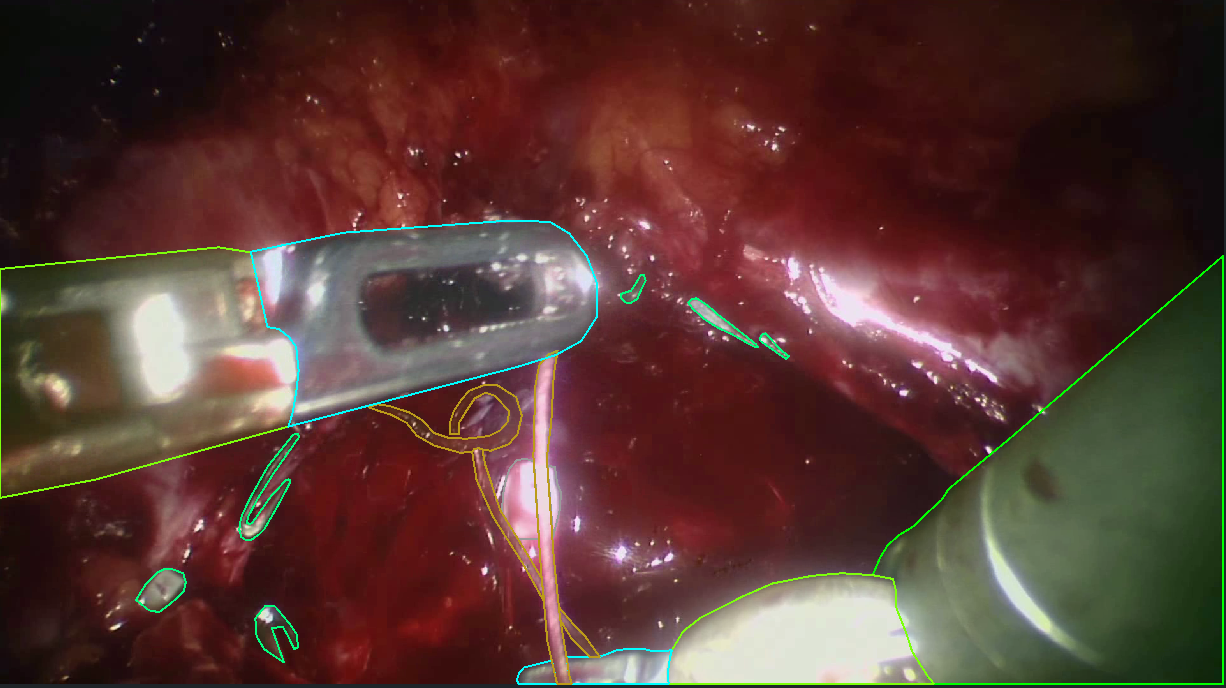}
      \caption{Labels of perforated tools tips}
      \label{fig:protocol_seg_2}
    \end{subfigure}\hfil 
    \begin{subfigure}{0.33\textwidth}
      \includegraphics[width=\linewidth]{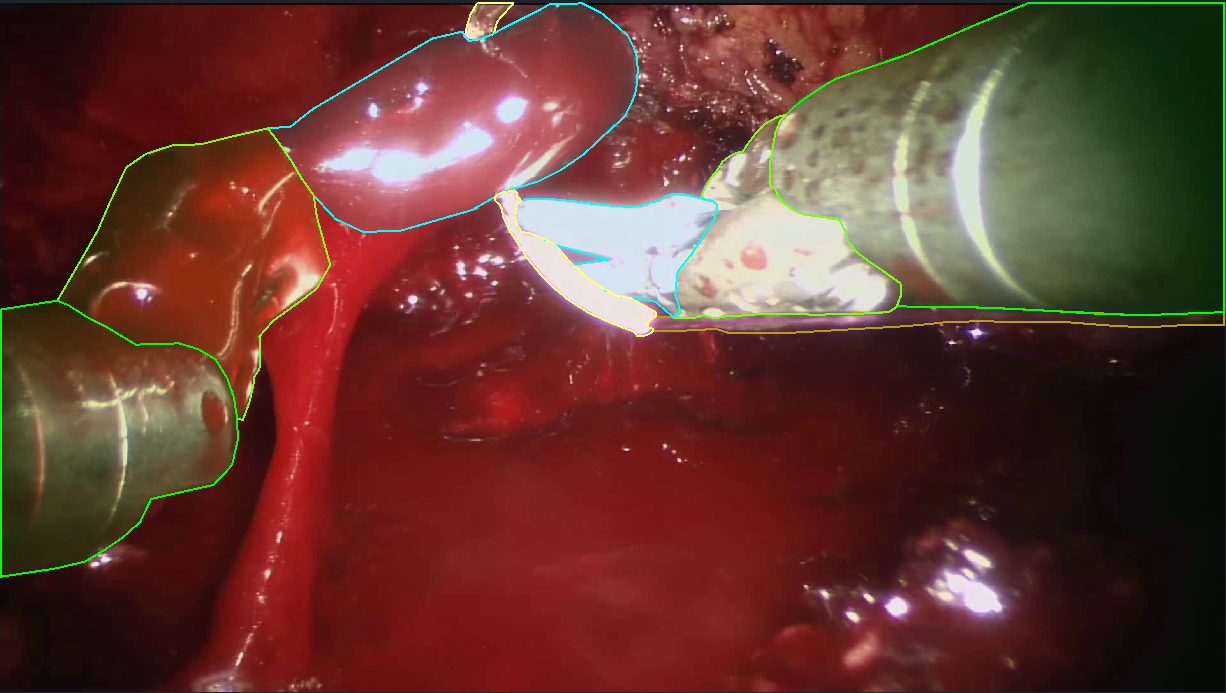}
      \caption{Labels of parts occluded by fluids}
      \label{fig:protocol_seg_3}
    \end{subfigure}

    \medskip
    \begin{subfigure}{0.33\textwidth}
      \includegraphics[width=\linewidth]{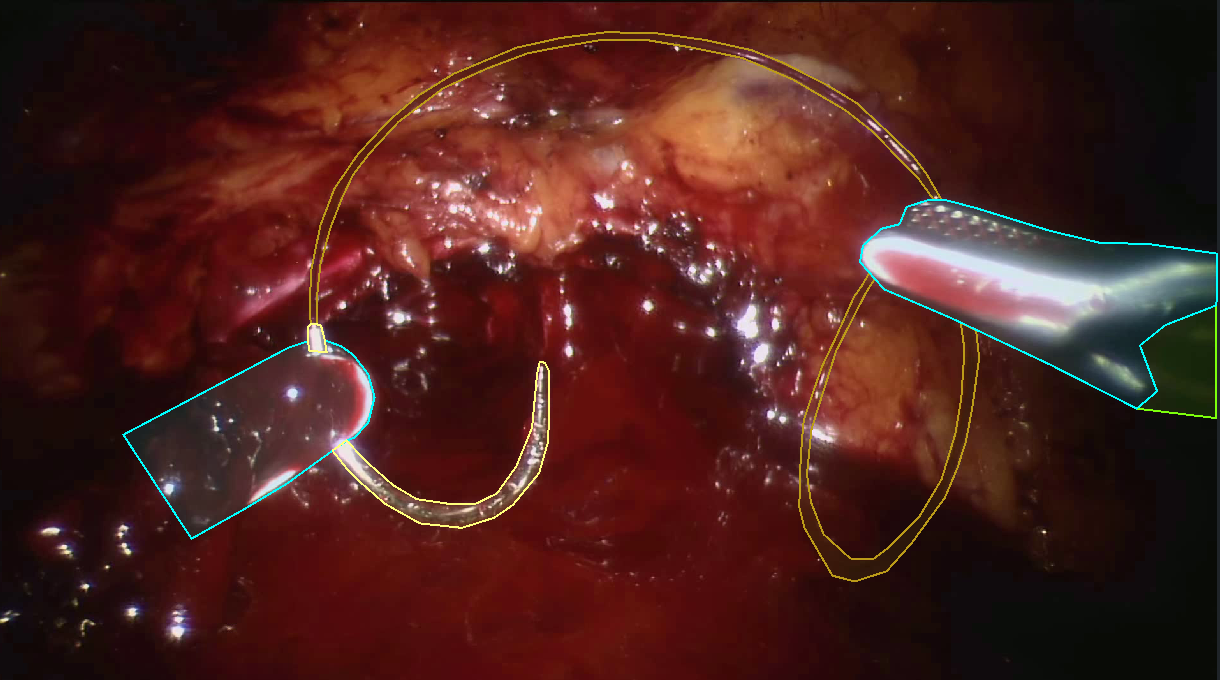}
      \caption{Labels of submerged objects}
      \label{fig:protocol_seg_4}
    \end{subfigure}\hfil 
    \begin{subfigure}{0.33\textwidth}
      \includegraphics[width=\linewidth]{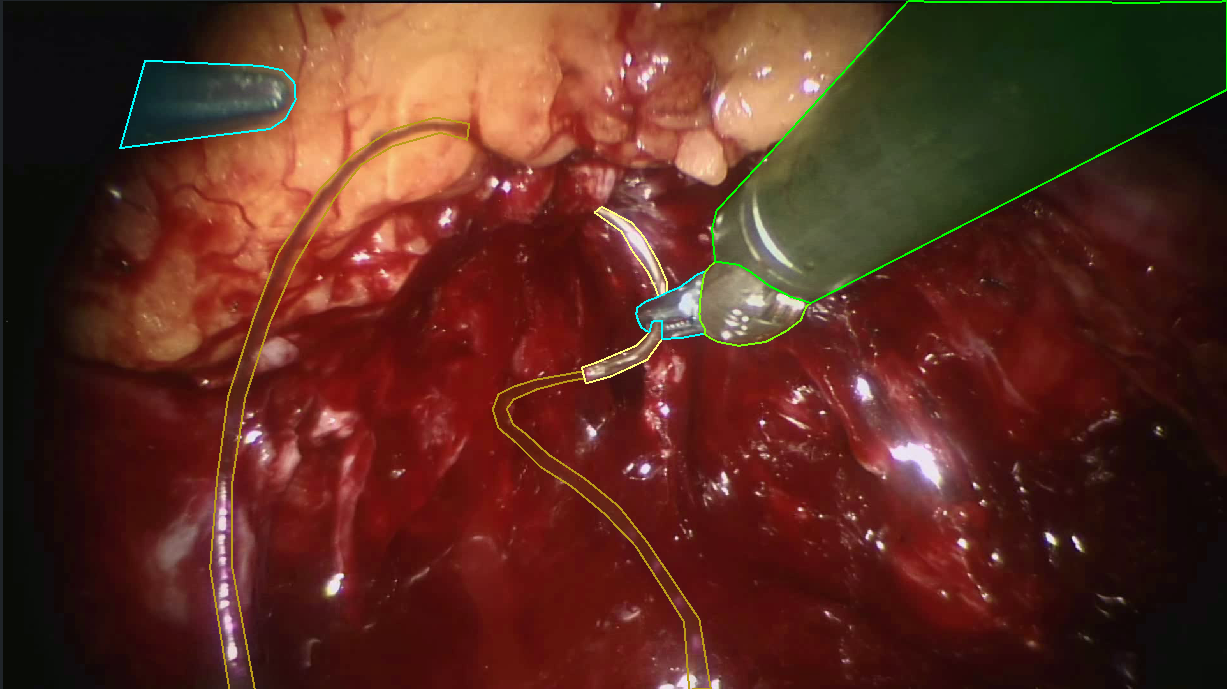}
      \caption{Label exclusion due to low illumination}
      \label{fig:protocol_seg_5}
    \end{subfigure}\hfil 
    \begin{subfigure}{0.33\textwidth}
      \includegraphics[width=\linewidth]{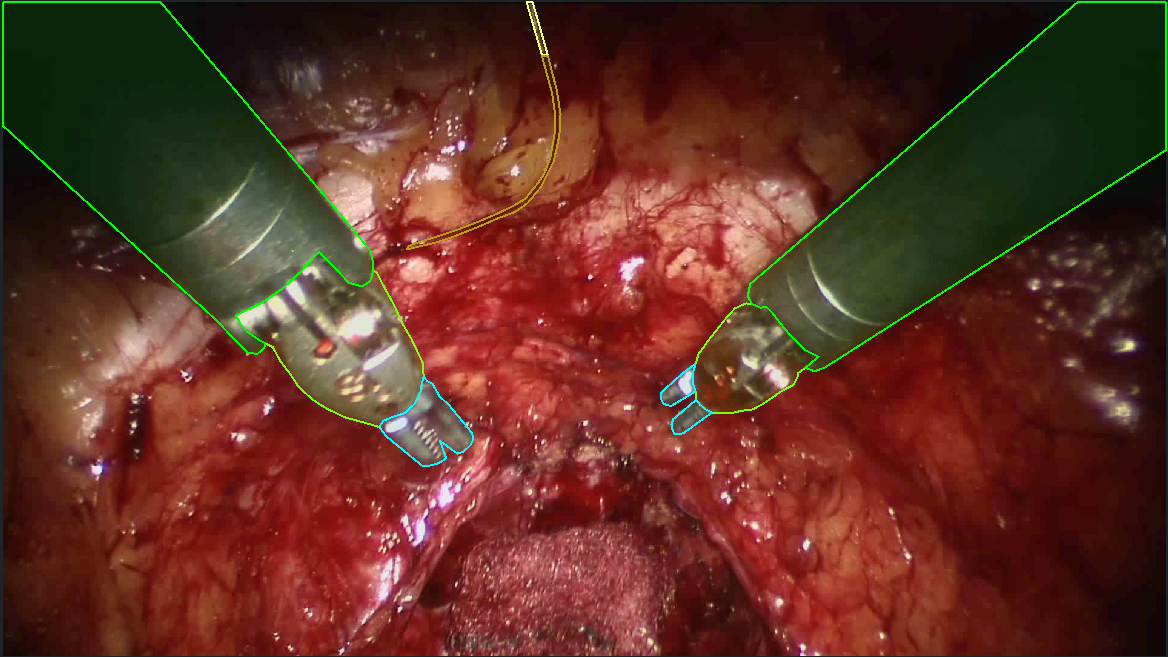}
      \caption{Propagating shaft labels}
      \label{fig:protocol_seg_6}
    \end{subfigure}
    \caption{Corner-cases of our semantic segmentation protocol to ensure consistent labels across SAR-RARP50.}
    \label{fig:images}
\end{figure*}

An annotation protocol is carefully designed to work with diverse operating conditions and edge cases captured across SAR-RARP50. The protocol aims at preserving context, even in cases where instruments are not clearly visible due to low illumination or when they are partially occluded by fluids or anatomy. The annotation protocol is defined as follows and resulting annotations are depicted in Fig.\ref{fig:images}:
\begin{enumerate}[a)]
    \item Each pixel can only correspond to a single semantic class. If objects from different semantic classes occlude each other, only the class of the object that occludes all the others is taken into account (Fig. \ref{fig:protocol_seg_1}) 
    \item Claspers that have holes, (i.e. Cadiere forceps and ProGrasp forceps), are labeled as if they were not perforated (Fig. \ref{fig:protocol_seg_2}).
    \item When fluids occlude surgical instrumentation by floating on top of or away from them, masks are defined to approximate the expected shape of the occluded object (Fig. \ref{fig:protocol_seg_3}). 
    \item Parts of instrumentation that are fully submerged in fluids are not annotated (Fig. \ref{fig:protocol_seg_4}).
    \item Tool parts near the edge of frames that are not clearly visible due to illumination, are not annotated (Fig. \ref{fig:protocol_seg_5}).
    \item Masks of tool shafts whose shape is clearly visible but fades towards the image edges due to vignetting, are extended until the edge of the frames (Fig. \ref{fig:protocol_seg_6}).
\end{enumerate}

The original video frames were sampled at a rate of 1Hz and uploaded to the Supervise.ly\footnote{https://supervisely.com/} online annotation platform. Humans In The Loop\footnote{https://humansintheloop.org/} (HITL) annotation service was tasked to create annotation masks for all frames following our annotation protocol. HITL assigned annotation generation to teams of expert annotators supervised by personnel with clinical expertise. Reference information generated by HITL was subsequently reviewed by two machine-learning researchers who assessed annotation quality and performed corrections. During the refinement process, labels were validated and refined based on information from the full video. 

\begin{figure}[!h]

    \centering 
    \includegraphics[width=\linewidth]{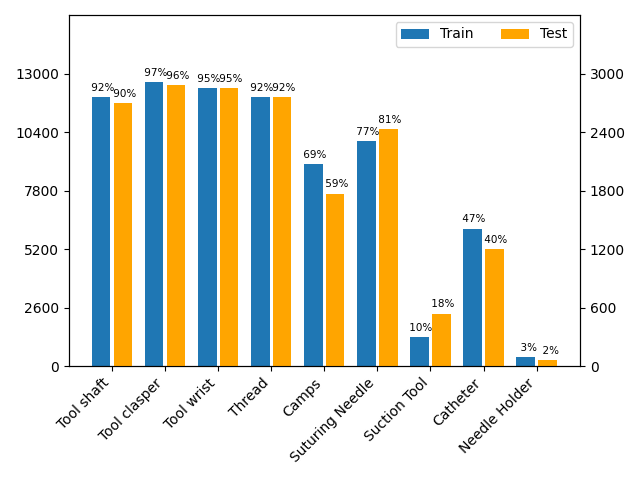}
    \caption{Segmentation class occurrence per sample in train and test sets. The y-axis corresponds to samples in the training set(left) and test set (right). Annotations show the percentage of samples depicting each class in the train set (blue) and test set (orange).}
    \label{fig:segmentation_class_occurance}
\end{figure}

The resulting dataset provides 12998 training frames from 40 different operations and 3252 test frames from 10 other operations. Fig.~\ref{fig:segmentation_class_occurance} shows the class occurrence among train and test sets. During the development of the dataset, operations were carefully assigned to test and training sets, such that the label distribution between the two sets was similar.

\begin{table}[t]
    \centering
    \caption{Average image coverage per class label across all samples of train and test set.}
    \begin{tabular}{|l|r|r|}
        \hline
        Class label     & Train set           & Test set             \\
        \hline
        Tool shaft      & $12.04 (\pm2.89)\%$   & $10.97 (\pm2.83)\%$    \\
        Tool clasper    & $2.24  (\pm1.19)\%$   & $3.61  (\pm2.13)\%$    \\
        Tool wrist      & $3.59  (\pm1.06)\%$   & $3.85  (\pm1.18)\%$    \\
        Thread          & $1.00  (\pm0.33)\%$   & $0.94  (\pm0.21)\%$    \\
        Clamps          & $0.15  (\pm0.19)\%$   & $0.13  (\pm0.16)\%$    \\
        Suturing needle & $0.44  (\pm0.15)\%$   & $0.45  (\pm0.15)\%$    \\
        Suction tool    & $0.51  (\pm0.53)\%$   & $0.70  (\pm1.03)\%$    \\
        Catheter        & $0.19  (\pm0.24)\%$   & $0.27  (\pm0.36)\%$    \\
        Needle Holder   & $0.20  (\pm0.15)\%$   & $0.23  (\pm0.12)\%$    \\  
        \hline
    \end{tabular}
        \label{tab:segmentation_class_area}
\end{table}

Table \ref{tab:segmentation_class_area} displays the average percentage of area covered by each class across the entire dataset. The provided values are computed across all samples. As a result, objects that occur less frequently such as the suction tools, exhibit low pixel coverage.

\section{Summary of submitted methods}
This section summarises  the methods proposed by each participating team in the SAR-RARP50 challenge. Participants varied in their submissions, with some exclusively contributing to a single sub-challenge, while three teams submitted solutions for all three sub-challenges.

\subsection{Team AIA-Noobs}

\textit{Semantic Segmentation}:
The team used an encoder-decoder CNN architecture consisting of an EfficientNetB4 \cite{tan2019efficientnet} encoder, pre-trained on Imagenet \cite{deng2009imagenet} and a UNet++ \cite{zhou2018unet++} decoder. This architecture was selected based on its accuracy from experiments the team conducted, combining different CNN feature encoders and decoders. They fine-tuned their model based on an equally weighted combination of Dice \cite{sudre2017generalised} and cross-entropy loss. During optimization, all RGB samples were resized to 480x640 resolution and normalized based on Imagenet colour statistics. Additionally, the team also applied +-15 degrees random rotations, +-10 degrees random shear and 90-100\% random scale augmentations. The proposed architecture was optimized for 100 epochs using the Ranger21 \cite{wright2021ranger21} optimizer with a learning rate of 1e-3 and momentum of 1e-9. The team used a data split of 7: 3 with a batch size of 16. To produce the final inference samples the group employed test time augmentations by averaging the predictions of their approach from the non-augmented and horizontally flipped frames.  

\textit{Multi-task}: 
The team proposed a network that used the predictions from their segmentation network as inputs to a ResNet18 \cite{he2015deep} feature extractor. Then, extracted features were used as input in a 2-layer LSTM \cite{hochreiter1997long} with 128 units responsible for task classification. This architecture was chosen based on the observation that action recognition based on segmentation masks yields more accurate predictions compared to RGB images. The proposed approach was jointly trained for both tasks. Action recognition was optimized based on the cross-entropy loss and the segmentation task, was trained based on a weighted sum of Dice \cite{sudre2017generalised} and cross-entropy loss. During training, the group used the provided target segmentation masks interpolated temporally from 1 to 10 Hz to match the sampling frequency of the action labels. The output of the segmentation network was fed directly to the action recognition sub-network without any reprocessing. The team optimized their approach using the same hyper-parameters as in their single-task segmentation network, except for using the Adam \cite{kingma2014adam} optimizer for the action recognition sub-network. To compute the final multitask output, the team performed test time augmentation for the segmentation task by horizontally flipping input samples and averaging the segmentation results.  

\subsection{Team CAMI-SIAT}
\textit{Action Recognition}: 
The team proposed combining Bridge-prompt \cite{li2022bridge} and ASFormer \cite{yi2021asformer} to infer per-frame action labels. This network combination was proposed because the semantic information contained in text labels provides richer context compared to the one-hot encoded reference labels. The group pre-trained Bridge-Prompt following the techniques described in ActionClip \cite{wang2021actionclip}. Optimization on SAR-RARP50 was done based on the weighted cross-entropy loss for ASFormer. Bridge-Prompt was optimized based on the KL divergence \cite{csiszar1975divergence} between the cosine similarity scores of the ground truth, and the outputs of the image and text encoders. The team used 28 videos for model training and performed random crops, colour jitter, horizontal flips and cutouts as data augmentation. Optimization was conducted for 50 epochs with a batch size of 96 using the AdamW \cite{loshchilov2017decoupled} optimizer and the cosine annealing \cite{loshchilov2016sgdr} learning scheduling policy with a warm-up of 5 epochs.  

\subsection{Team HiLab-2022}

\textit{Semantic Segmentation}: 
The team proposed using an ensemble of a Swin Transformer \cite{liu2021swin} and a SegFormer \cite{xie2021segformer} to predict segmentation masks. To optimize the two networks, the group created a training set by randomly sampling 10824 images from SAR-RARP50 dataset resized to 1920x512 resolution. They further pre-process and augment their training data by applying colour normalization, random scale, random flips and cropping frames to 512x512 resolution. Both networks were optimized based on the cross-entropy loss using stochastic gradient descent (SGD) with momentum of 0.9, learning rate of 1e-2, gamma of 0.5, weight decay of 5e-4 and a batch size of 4. The Swin transformer was optimized from scratch while the optimization of SegFormer started from a pre-trained version of the network. To produce the final results, the group performed test time augmentation by flipping the input images and averaging the output of all networks, using the largest K-th connected domain to retrieve the final result. 

\subsection{Team Kings-SurgicalAI}
\textit{Action Recognition}: 
This team proposed a modification of their previous work LoViT - Long Video Transformer \cite{liu2023lovit} for action recognition. They use a Vision Transformer (ViT) \cite{dosovitskiy2020image} to extract spatial features from all frames in short video sequences. Next, the extracted spatial embeddings are aggregated temporally by forward and backward local-temporal feature aggregator transformer modules(L-Trans). In the forward branch, L-trans modules accept as input the features of the current frames and the output of L-trans making predictions for the previous frame in time. In contrast, L-trans modules of the backward branch accept features of the current frame and the output of L-trans of the following frame in sequence. To predict an action for a frame, they use a global Informer \cite{zhou2021informer} with ProbSparse self-attention with inputs the sum of all L-Trans outputs and the output of L-Trans associated with the video sequence starting from the query frames. 

\subsection{Team Medical-Mechatronics}
\textit{Action Recognition}: 
The team proposed to make action recognition predictions using Xception \cite{chollet2016xception}, a model based on depthwise separable convolution layers. The proposed model was pre-trained on Imagenet \cite{deng2009imagenet} and then fine-tuned on SAR-RARP50 based on the cross-entropy loss. The group used dataset 1-4, 6, 7, 9, 10, 11\_2, 13,15\_1, 17\_1, 18, 20-28, 29\_2, 30-34, 36-40 for training and reserved the rest of the videos for evaluation.  Optimization on the target dataset was done using the Adam \cite{kingma2014adam} optimizer with an initial learning rate of 1e-3 and a batch size of 64. During optimization, the learning rate was reduced at a rate of 0.957 every 10 epochs. 

\textit{Semantic Segmentation}: 
They proposed to solve the surgical instrument segmentation task using the LinkNet architecture \cite{chaurasia2017linknet} with ResNet-34 \cite{he2015deep} encoder. The team chose this combination because it has been known to perform well when fine-tuned to the surgical domain \cite{shvets2018automatic}. They optimized their approach based on a weighted combination of the Intersection-over-Union (IoU) and negative log-likelihood (NLL) loss functions for 100 epochs using the Adam \cite{kingma2014adam} optimizer with a learning rate of 1e-4 and a batch size of 64. The group trained and evaluated their approach using the same data split as in their action recognition submission. During optimization, the team resized training samples to 224x244 and applied random horizontal and vertical flips augmentation with a probability of 50\%. 

\subsection{Team NCC-Next}
\textit{Action Recognition}: 
The team proposed to predict actions using two models, a Video Swin transformer \cite{liu2022video} and SlowFast \cite{feichtenhofer2019slowfast}. The two models were trained using 5-fold cross-validation, resulting in an ensemble of 10 models to predict an action label. This approach was chosen as it combines predictions from both transformer and convolutional architectures. The SlowFast network was implemented with a re-sampling rate to a slow path of 2, a sampling rate between fast and slow pathways of 2, and a sampling rate between fast and slow pathways of 8. Both SlowFast and the Video Swin transformer were pre-trained on Kinetics400 \cite{kay2017kinetics} and optimized on SAR-RARP50 for 10 epochs based on the cross-entropy loss using a batch size of 16. During training, image samples were normalized based on the SAR-RARP50 colour statistics and resized to 256x512 resolution. Data augmentations included perspective transformations, colour adjustments and cutouts. The video Swin transformer was optimized using the AdamW \cite{loshchilov2017decoupled} with an initial learning rate of 0.01 and weight decay of 0.02. SlowFast was optimized using SGD with momentum of 0.8, learning rate of 0.1 and weight decay of 10e-4. During the optimization of both architectures, the cosine annealing \cite{loshchilov2016sgdr} learning rate policy with a warm-up period of 5 epochs was used. During testing, initial network predictions were further processed based on a multi-scale filtering mechanism after which, the predictions of all networks were averaged to produce the final result.  

\textit{Semantic Segmentation}: 
The team formed an ensemble of 3 different architectures to estimate tool segmentation masks. The first model used a Swin transformer \cite{liu2021swin} as an encoder and a UperNet \cite{xiong2019upsnet} as a decoder. The second model used an HRNetV2 \cite{sun2019high} as an encoder and an OCRNet \cite{yuan2020object} as a decoder. The last model used a MiT-B3 encoder and a SegFormer decoder \cite{xie2021segformer}. The encoders of all models were pre-trained on Imagenet \cite{deng2009imagenet}. Optimization on SAR-RARP50 was done for 30 epochs using the RAdam \cite{liu2019variance} with a learning rate of 5e-5 modified based on cosine annealing \cite{loshchilov2016sgdr} and warmup policies. During training the team applied random image flips, shifts, scaling, and rotations to augment the dataset. The first two models were optimized based on the sum of focal(gamma=2) and Dice loss with a batch size of 32 and weight decay of 1e-5. The third model was optimized on the cross entropy loss with a batch size of 9 and weight decay of 1e-4. During testing, segmentation masks of all models were aggregated using channel-wise mask summation followed by an Argmax operation.

\subsection{Team Orsi-Academy}
\textit{Semantic Segmentation}: 
This team proposed to use a Feature Pyramid Network (FPN) \cite{lin2017feature} based on the EfficientNetV2-S \cite{tan2021efficientnetv2} backbone because of its excellent run-time performance. They allocated 37 SAR-RARP50 videos for training and evaluated their method on the remaining 7. They optimized their network from scratch on SAR-RARP50 based on the Focal loss, using AdamW \cite{loshchilov2017decoupled} with learning rate of 2e-4 and a batch size of 8 samples. During training, the group reduced the learning rate when loss plateaued and implemented early stopping with patience of 5 epochs. The team augmented the training data by applying horizontal flips, rotation, brightness, and contrast modification and adding Gaussian noise and motion blur. 

\subsection{Team SK}
\textit{Multi-task}: 
Team SK proposed a multi-task network consisting of a shared, among tasks, ResNet-101 \cite{he2015deep} encoder with a U-Net++ \cite{zhou2018unet++} head responsible for predicting segmentation masks and a fully connected layer as a second head, dedicated to action recognition predictions. The team selected this architecture based on accuracy, after evaluating all combinations of ResNet-50 and ResNet-101 with U-Net \cite{ronneberger2015u} and Unet++. The proposed approach was optimized end-to-end based on an equally weighted combination of t-vMF Dice loss \cite{kato2022adaptive} for segmentation results and the cross-entropy loss for the action recognition predictions. The team used samples from 38 videos sampled at 1 FPS and resized to 512x512 resolution. During training the team, applied horizontal flip, $\pm5$ spatial shift, $\pm5$ scaling, $\pm5\deg$ Rotation, $\pm5\%$ colour jitter, Gaussian blur (sigma is 3 to 7), and Gaussian noise ($\sigma \in [10, 50]$) data augmentations with probability of 0.5. Optimization was performed for 30 epochs and a batch size of 16, using the AdamW \cite{loshchilov2017decoupled} optimizer with an initial learning rate of 1e-4 and weight decay of 1e-5. During training the learning rate was modified based on the cosine annealing policy \cite{loshchilov2016sgdr}.  

\subsection{Team SummerLab-AI}
\textit{Action Recognition}: 
The team proposed using an ASFormer \cite{yi2021asformer} model, to model the temporal relationship among frames and predict per-frame action labels. In their approach, instead of feeding a vanilla ASFormer with raw RGB frames, they optimized a Bridge-prompt \cite{li2022bridge}, of which the image feature encoder was used to pre-extract frame-wise features to serve as inputs of ASFormer. The group selected this approach for its inherent local inductive bias and effective representation of long input sequences. The Bridge-Prompt model was pre-trained on Kinetics400 \cite{kay2017kinetics}. During training on SAR-RARP50, the team resized all samples to 422x750 and applied random colour jitter, random horizontal flips, and random grayscale data augmentations. Fine-tuning was done based on the sum of a per-frame cross-entropy loss and a smooth loss. The smooth loss was weighted by 0.25 and computed as the mean squared error over the frame-wise probabilities. The proposed approach was developed using a 5-fold cross-validation scheme with all networks optimized for 100 epochs using the Adam optimizer with a batch size of 20. The initial learning rate was 1e-5, and modified during training using the cosine annealing policy \cite{loshchilov2017decoupled}. 

\textit{Semantic Segmentation}: 
The group proposed using Swin Transformer Large \cite{liu2021swin} feature encoder with an UperNet \cite{xiong2019upsnet} decoder to generate the final segmentation masks. This architecture was chosen as it performed the best against different encoder and decoder combinations the team tested on SAR-RARP50. The team fine-tuned their approach for SAR-RARP50 using 5-fold validation, starting with a pre-trained encoder on ImageNet-22K \cite{deng2009imagenet}. Optimization was conducted based on a combination of  cross-entropy Loss and Dice Loss waited with 0.75 and 0.25, respectively. During training, the team resized all samples to 422x750 and applied Random brightness, contrast, motion blur and horizontal flip augmentation to increase the model's performance under the lighting conditions present on SAR-RARP50. Optimization on SAR-RARP50 was performed using AdamW \cite{loshchilov2017decoupled} with a batch size of 4. The initial learning rate was set to 6e-5 and modified during training based on the cosine annealing \cite{loshchilov2016sgdr} policy. To generate the final predictions, the group used multi-scale (0.5, 0.75, 1, 1.25, 1.5) and horizontal flips test time augmentation. 

\textit{Multi-task}: 
The team inspired by the Multi-Task Recurrent Convolutional Network \cite{jin2020multi}, proposed a multitask architecture that extends their single-task segmentation model with a Bi-LSTM, processing features from the common Swin Transformer \cite{liu2021swin} backbone, to recognise action. The team did not opt to use components from their single-task action recognition model due to time constraints. The proposed approach was optimized jointly using the cross-entropy loss for the action recognition and a sum of the cross-entropy and dice loss functions weighted by 0.75 and 0.25 respectively for semantic segmentation. The group trained their model using 1Hz video samples, resized to 256x256 resolution. They trained with a batch size of 20 samples using the AdamW \cite{loshchilov2017decoupled} optimizer with an initial learning rate of 6e-5 modified by a cosine annealing scheduler \cite{loshchilov2016sgdr}. During inference, their network was fed with 10 consecutive frames allowing the Bi-LSTM to make predictions for every 2 consecutive frames. The team performed a soft ensemble, aggregating multiple action recognition predictions for the same frame.  

\subsection{Team TheOne-Lab}
\textit{Semantic Segmentation}: 
This team proposed to use a modified version of HRNet \cite{sun2019high} to solve the tool segmentation task. This model was chosen because it links high and low-resolution features in parallel, allowing it to produce detailed segmentation masks while maintaining good geometry characteristics of the target classes. The group modified HRNet by increasing its depth and receptive field, improving the quality of the features the network can normally extract. The model was optimized on SAR-RARP50, without any pre-training step, based on the cross-entropy loss. The team split SAR-RARP50 intro training and validation using a 10:1 ratio and applied random scaling and flipping augmentation techniques. Optimization was conducted with a batch size of 2, for 25 epochs, using the SGD optimizer with a learning rate of 0.01, momentum of 0.9 and weight decay of 5e-4. During training, the learning rate was updated based on lambdaLR and a 5-epoch warmup policy. To generate test predictions, the team employed test time augmentations by merging network predictions for the same sample inferred at 1, 0.83 and 0.67 scales. 

\subsection{Team TSO22}

\textit{Action Recognition}: 
The team propose to solve the problem in two stages. First, they used a ResNet50 \cite{he2015deep} to extract features for every frame individually. Second, to process the extracted features, they used a 2-stage MS-TCN \cite{farha2019ms} with 10 temporal convolution layers per stage and 64 feature maps. Their framework was trained in two stages. The ResNet 50 was optimized for 50 epochs using the AdamW \cite{loshchilov2017decoupled} optimizer with a learning rate of 1e-4, based on a weighted cross-entropy loss. They pre-processed all input images by first resizing them to 640x360, cropping them to 324x324 and applying geometric and colour augmentations. In the second stage, they used a ResNet50 to pre-compute features for all images, to later use as inputs to MS-TCN. They trained MS-TCN for 200 epochs using AdamW with a learning rate of 1e-4 based on the cross-entropy loss. 

\textit{Semantic Segmentation}: 
The team proposed using a SegFormer-B1 \cite{xie2021segformer} to make segmentation predictions. They used 34 SAR-RARP50 videos 
for training and allocated the rest for evaluation. Starting with a pre-trained encoder on Imagenet \cite{deng2009imagenet}, the group fine-tuned their model on SAR-RARP50 for 100 epochs. The team augmented the provided dataset by applying random shift, scaling, rotation, colour noise, brightness, and contrast perturbations. Optimization was done using the AdamW \cite{loshchilov2017decoupled} optimizer with a learning rate of 6e-5 and a batch size of 2, based on the cross-entropy loss.

\subsection{Team Uniandes}

\textit{Action Recognition}: 
This team proposed TAPIR, an architecture using an MViT \cite{fan2021multiscale} to encode video information and a single-layer MLP classifier to perform action recognition. The feature encoder processes a 32-frame sequence, sampled using a stride of 12 and centred on the target frame. The group used a pre-trained backbone on Kintetics400 \cite{kay2017kinetics} for short video clip classification and then finetuned their whole network on SAR-RARP50, based on the cross-entropy loss. The team trained their approach on datasets 1, 3, 7, 13, 14, 18, 20-40 and validated with the rest of the released training set. During training, all frames of a window sequence were augmented the same way, by applying, random horizontal flips, random resizing maintaining the aspect ratio, and finally random cropping to 224x224. Optimization was conducted for 15 epochs, with a batch size of 9 time windows, each centred on the target frame,  using SGD with a base learning rate of 1,25e-2, momentum of 0.9 and weight decay of 1e-8. During training, the learning rate was modified based on the cosine annealing policy \cite{loshchilov2016sgdr} and a warm-up period 5 epochs. During inference, initial predictions for the whole sequences were further processed by a 250-step iterative filtering mechanism. This filter replaced the class of low-scoring action-transition frames with the class of the highest-scoring neighbour.  

\textit{Semantic Segmentation}: 
The team proposed to solve the tool segmentation task using Mask2Former \cite{cheng2022masked} with a Swin Base \cite{liu2021swin} backbone. Since the challenge day, the team has published the presented approach \cite{ayobi2023matis}. This method was chosen because it uses a mask classification approach, which according to the team, is better than the conventional pixel-level approach in segmenting tools. Additionally, the region proposal functionality of Mask2Former could serve as a building block for a multi-task architecture, helping the action recognition task. Starting from Swin B mask2Former trained on COCO \cite{lin2014microsoft} for instance segmentation, they first fine-tuned their model on Endovis 2017 \cite{allan20192017} and Endovis 2018 \cite{allan20202018} datasets and then fine-tuned their approach on SAR-RARP50 using the same data split as in their action recognition approach. They used the same optimization criterion as in the initial Mask2Former paper. During training, they resized all training samples to 750x1333 and performed random horizontal flips. Optimization was done using the AdamW \cite{loshchilov2017decoupled} optimizer with a weight decay of 0.5, a multi-step learning rate scheduler starting with at 1e-4  decaying by 0.1 at epochs 44, 48, 66, and 72. The training was completed after 75 epochs with a batch size of 16. To make final predictions, the team selected the top 10 scoring binary masks, among the 100 inferred masks Mask2Former, per frame, and later discarded masks with a score lower than 0.75. Finally, they chose the highest-scoring mask for pixels segmented in more than one binary mask.

\textit{Multi-task}: 
Uniandes proposed to combine the architectures they submitted in the single-task sub-challenges into one. Their approach makes predictions based on a time window centred on the target frame. Gesture predictions were based on the class embedding output by MViT \cite{fan2021multiscale}. This embedding is computed from the input time window sequence and a learnable class token. The semantic segmentation output is predicted on the remaining spatiotemporal embeddings and the pre-computed regions proposals of the middle frame. The team proposed this solution as it was a natural extension of their single-task submissions. The resulting architecture was novel and based on previous work  of the Unianders team members \cite{valderrama2022towards}. Starting from their SAR-RARP50 fine-tuned models, the group trained their approach relying on the single-task pre-computed region proposal and features from their segmentation model and optimized the remaining components based on the cross-entropy loss of each task. Optimization was similar to the action recognition task, except it lasted 25 epochs.

\section{Results and Discussion}

\subsection{Surgical action recognition}

The results for the action recognition sub-challenge are presented in Table~\ref{table:res_ar_f1} and Table~\ref{table:res_ar_acc}, in terms of $F@10$ score and accuracy, respectively. The final ranking is equal for both evaluation metrics, with team Summerlab-AI at the lead with a final score of 0.82, followed by Uniandes with 0.80 and CAMI-SIAT with 0.78 (Table \ref{table:res_ar_final}). A final stability study computed based on eq. (\ref{loss:var}), and presented in the Table~\ref{table:res_ar_rank}, further confirms the ranking. 

\begin{table*}[bth!]
    \centering
    
    \caption{Per Video F1@10~($\uparrow$) action recognition score. Bold indicates the highest score among teams.}
    \begin{tabular}{l|cccccccccc}
        Team                & 41    & 42    & 43    & 44    & 45    & 46    & 47    & 48    & 49    & 50    \\
        \hline
        SummerLab-AI         & \textbf{0.921} & \textbf{0.845} & \textbf{0.831} & \textbf{0.980} & 0.926 & 0.667 & 0.771 & \textbf{0.932} & \textbf{0.715} & 0.813 \\
        Uniandes             & 0.892 & 0.708 & 0.811 & 0.943 & \textbf{0.963} & \textbf{0.696} & 0.756 & 0.879 & 0.693 & \textbf{0.889} \\
        CAMI-SIAT            & 0.831 & 0.780 & 0.822 & \textbf{0.980} & 0.943 & 0.548 & 0.771 & 0.895 & 0.651 & 0.842 \\
        NCC-Next             & 0.885 & 0.717 & 0.812 & 0.863 & 0.824 & 0.675 & \textbf{0.853} & 0.844 & 0.671 & 0.843 \\
        TSO22                & 0.676 & 0.722 & 0.773 & 0.820 & 0.877 & 0.533 & 0.594 & 0.816 & 0.556 & 0.707 \\
        KingSurgial-AI       & 0.532 & 0.246 & 0.455 & 0.511 & 0.539 & 0.409 & 0.312 & 0.548 & 0.304 & 0.446 \\
        Medical-Mechatronics & 0.000 & 0.000 & 0.051 & 0.077 & 0.000 & 0.000 & 0.000 & 0.000 & 0.000 & 0.000
    \end{tabular}
    \label{table:res_ar_f1}
\end{table*}

\begin{table*}[bth!]
\centering
\caption{Per Video Accuracy~($\uparrow$) action recognition score. Bold indicates the highest score among teams.}
\begin{tabular}{l|cccccccccc}
Team                & 41    & 42    & 43    & 44    & 45    & 46    & 47    & 48    & 49    & 50    \\
\hline
SummerLab-AI         & \textbf{0.826} & \textbf{0.748} & \textbf{0.859} & \textbf{0.917} & \textbf{0.863} & 0.690 & \textbf{0.842} & \textbf{0.893} & 0.733 & 0.781 \\
Uniandes             & 0.821 & 0.650 & 0.800 & 0.873 & 0.837 & \textbf{0.777} & 0.771 & 0.771 & \textbf{0.741} & 0.813 \\
CAMI-SIAT            & 0.789 & 0.719 & 0.763 & 0.860 & 0.861 & 0.598 & 0.752 & 0.838 & 0.702 & \textbf{0.818} \\
NCC-Next             & 0.777 & 0.690 & 0.669 & 0.659 & 0.709 & 0.715 & 0.767 & 0.700 & 0.720 & 0.722 \\
TSO22                & 0.663 & 0.697 & 0.732 & 0.754 & 0.768 & 0.579 & 0.677 & 0.734 & 0.649 & 0.643 \\
KingSurgial-AI       & 0.649 & 0.465 & 0.637 & 0.681 & 0.676 & 0.670 & 0.493 & 0.593 & 0.521 & 0.599 \\
Medical-Mechatronics & 0.077 & 0.069 & 0.225 & 0.154 & 0.073 & 0.178 & 0.103 & 0.083 & 0.056 & 0.153
\end{tabular}
\label{table:res_ar_acc}
\end{table*}

\begin{table*}[bth!]
        \centering
        \caption{Per video action recognition ranking stability, computed based on eq. \ref{loss:var}}
        \begin{tabular}{l|cccccccccc|c}
            Team                & 41 & 42 & 43 & 44 & 45 & 46 & 47 & 48 & 49 & 50 & Average~($\uparrow$) \\
            \hline
            SummerLab-AI         & 1  & 1  & 1  & 1  & 3  & 3  & 2  & 1  & 1  & 3  & 1.7     \\
            Uniandes             & 2  & 5  & 2  & 3  & 2  & 1  & 3  & 3  & 2  & 1  & 2.4     \\
            CAMI-SIAT            & 4  & 2  & 3  & 2  & 1  & 4  & 4  & 2  & 4  & 2  & 2.8     \\
            NCC-Next             & 3  & 4  & 5  & 5  & 5  & 2  & 1  & 5  & 3  & 4  & 3.7     \\
            TSO22                & 5  & 3  & 4  & 4  & 4  & 5  & 5  & 4  & 5  & 5  & 4.4     \\
            KingSurgial-AI       & 6  & 6  & 6  & 6  & 6  & 6  & 6  & 6  & 6  & 6  & 6       \\
            Medical-Mechatronics & 7  & 7  & 7  & 7  & 7  & 7  & 7  & 7  & 7  & 7  & 7      
        \end{tabular}
        \label{table:res_ar_rank}
\end{table*}

\begin{table}[bth!]
    \centering
    \caption{Aggregated results. The first and second columns represent the average Accuracy and F1@10 scores across the test set respectively. Teams were ranked based on the Final score computed according to eq. ~\ref{loss:ar}}
    \resizebox{0.47\textwidth}{!}{
        \begin{tabular}{l|cc|c}
            Team                 & Accuracy~($\uparrow$) & F1@10~($\uparrow$)  & Final Score~($\uparrow$) \\
        \hline
            SummerLab-AI         & \textbf{0.815}    & \textbf{0.841}  & \textbf{0.828} \\
            Uniandes             & 0.786    & 0.823  & 0.804 \\
            CAMI-SIAT            & 0.770    & 0.806  & 0.788 \\
            NCC-Next             & 0.713    & 0.799  & 0.755 \\
            TSO22                & 0.690    & 0.707  & 0.698 \\
            KingSurgial-AI       & 0.598    & 0.430  & 0.507 \\
            Medical-Mechatronics & 0.117    & 0.013  & 0.039
        \end{tabular}
        }
    \label{table:res_ar_final}
\end{table}
    

The solutions proposed by participants can be broadly categorized into two types: 
\begin{itemize}
    \item two-stage approaches involving a feature extractor followed by a long-range temporal model
    \item single-stage window-based approaches with post-processing for prediction smoothing.
\end{itemize}
Overall, the two-stage approaches performed better due to their capacity to process long temporal sequences at once. The top-performing methods across the challenge were predominantly attention-based models, highlighting the effectiveness of this architecture in the domain of surgical action recognition. The superiority of attention-based models was also confirmed by CAMI-SIAT who while developing their methods tested different combinations of CNN and transformer-based models for their two-stage approach before selecting their fully attention-based architecture.
Approaches pre-trained on large video action datasets \cite{kay2017kinetics} performed significantly better compared to those that did not. SummerLab-AI and CAMI-SIAT propose similar network architectures but different optimization criteria and pre-training, resulting in significant differences in scores. Another interesting insight regarding post-processing techniques was provided by Uniandes. They tested both classic and learning-based filtering methods to filter action recognition predictions and found that classic window-based filtering performed the best. 

\begin{figure*}[bth!]
    \centering
    \includegraphics[width=0.8\textwidth, center]{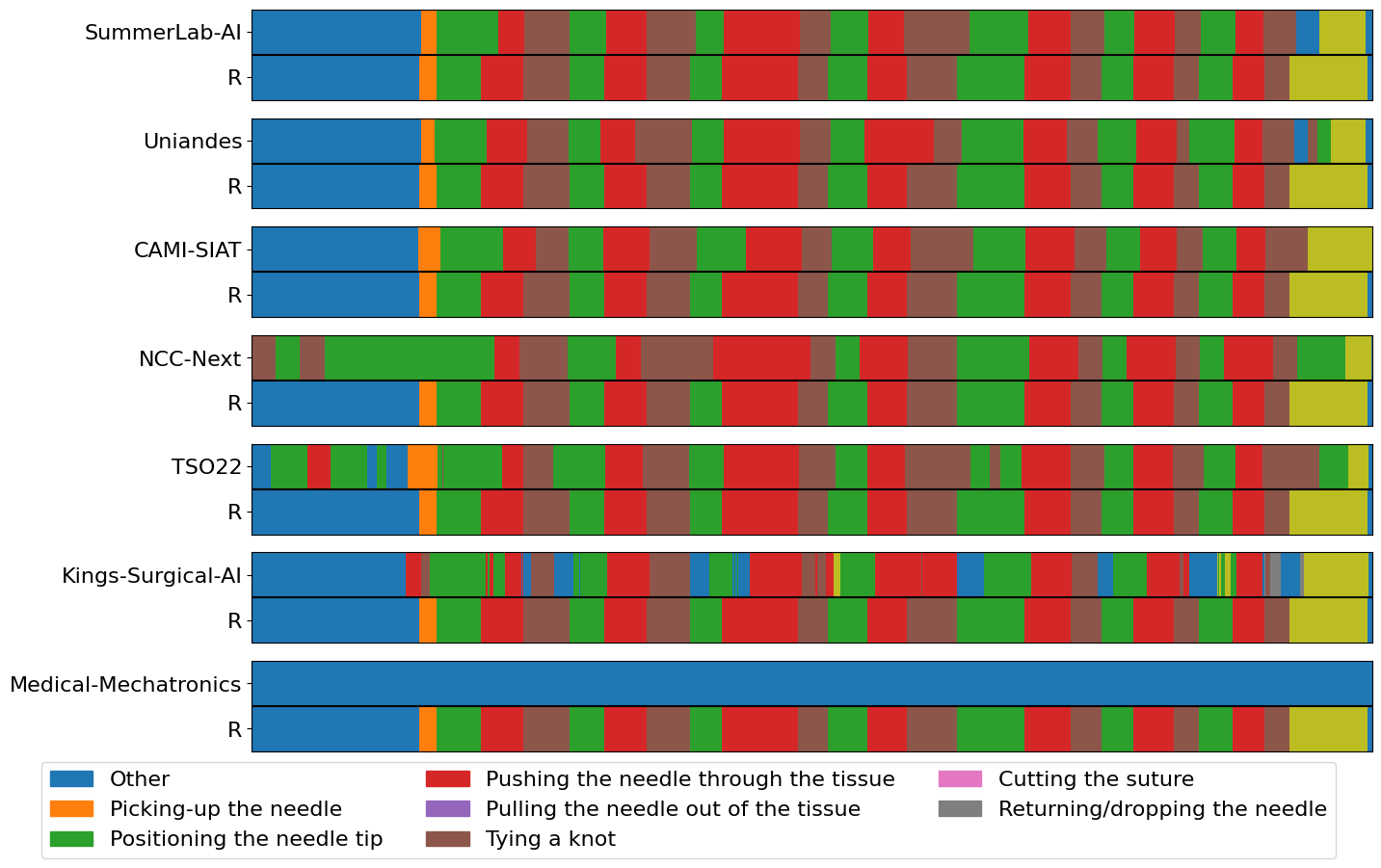}
    \caption{Timeseries action graph for Video 44 which corresponds to an operation performed by an expert surgeon. The upper segment of each box corresponds to method predictions, while R stands for reference.}
    \label{fig:res_ar_1}
\end{figure*}

\begin{figure*}[bth!]
    \centering
    \includegraphics[width=0.8\textwidth, center]{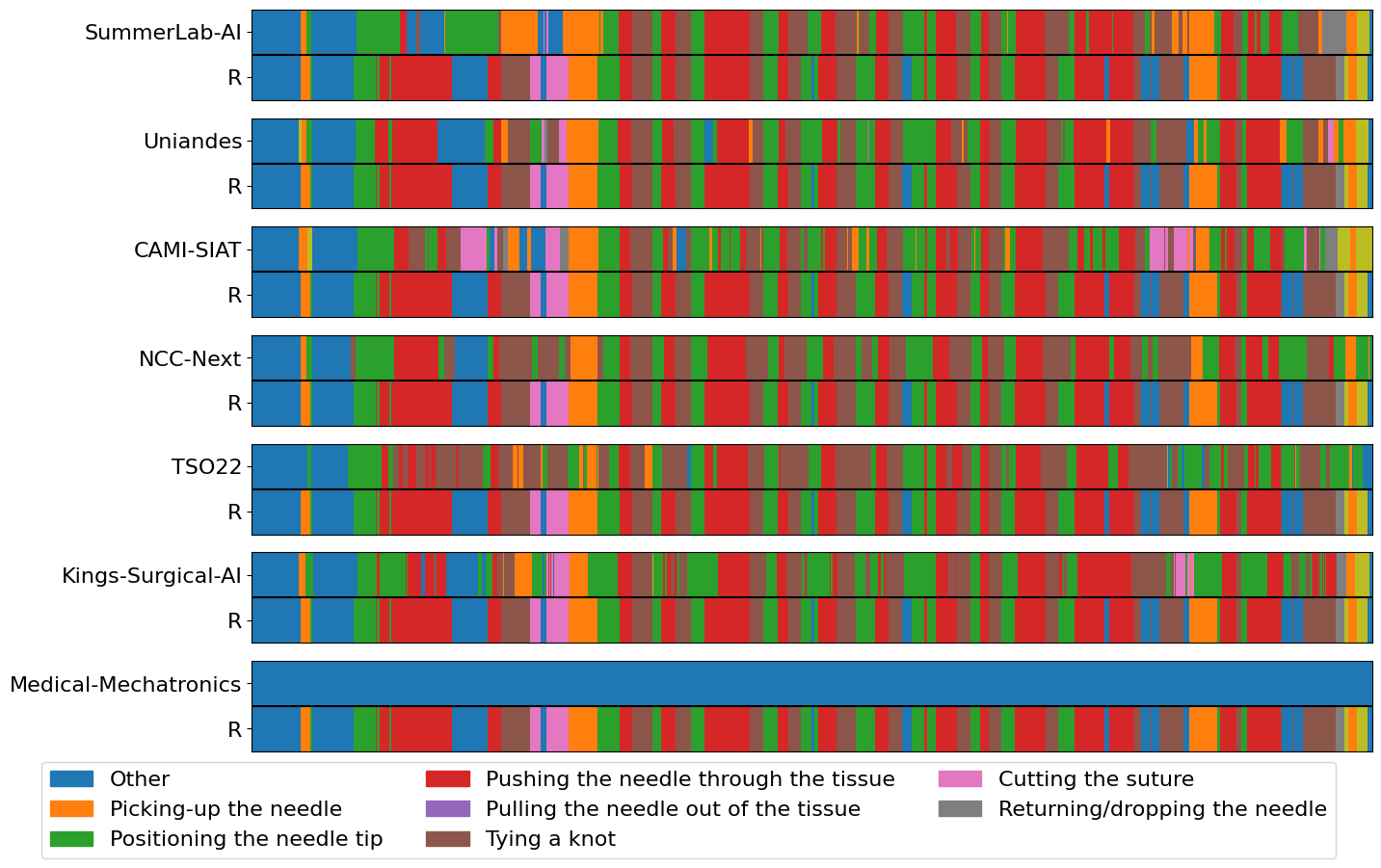}
    \caption{Timeseries action graph for Video 46 which corresponds to an operation performed by a junior registrar. The upper segment of each box corresponds to method predictions, while R stands for reference.}
    \label{fig:res_ar_2}
\end{figure*}

The test sequence which yields the highest recognition scores across all methods is illustrated in Fig.\ref{fig:res_ar_1} and corresponds to a surgery conducted by an experienced consultant. This sequence exhibits minimal bleeding, limited camera motion, minimal assistant intervention, and a fairly regular series of gestures. On the contrary, the test sequence with the lowest recognition scores, shown in Fig.\ref{fig:res_ar_2}, corresponds to a surgery performed by a junior registrar. This sequence involves more bleeding, camera motion, assistant intervention and a longer and less regular gesture sequence. These findings suggest that incorporating videos from junior surgeons or challenging interventions can be beneficial for improving the robustness of recognition models. Additionally, the integration of real surgical videos and surgical training data can positively contribute to enhanced performance.

\subsection{Surgical Instrumentation Semantic Segmentation}

Tables ~\ref{table:seg_iou} and ~\ref{table:seg_nsd} show the per-video segmentation performance for every team based on mIoU and mNSD respectively. Aggregated results across all test sequences for each metric and the final score are presented in Table~\ref{table:seg_all}. Overall, the top 3 teams demonstrated very similar performance. Team Uniandes achieved the best overall segmentation score of 0,847, followed by HiLab-2022 in second place who scored 0.840, and SummerLab-AI scoring 0.839 in third place. Team ranking is the same in both the final score and individual metrics, however when looking at the average per video ranking presented in table \ref{table:seg_rank}, we see that SumerLab-AI comes second and HiLab-2022 becomes third. This change in ranking is interesting and demonstrates how using different aggregation techniques and metrics can have a major impact on validation, as described in~ \cite{maier2018rankings}.

\begin{table*}[bth!]
\centering
\caption{Per-video segmentation mIoU~($\uparrow$) results. Scores highlighted in bold show the top-performing method for each test video.}
\begin{tabular}{l|cccccccccc}
Team                & 41             & 42    & 43    & 44    & 45    & 46    & 47    & 48    & 49    & 50    \\
\hline
Uniandes             & \textbf{0.862} & \textbf{0.810} & \textbf{0.843} & \textbf{0.798} & \textbf{0.839} & \textbf{0.817} &\textbf{0.825} & \textbf{0.869} & \textbf{0.797} & \textbf{0.833} \\
HiLab-2022           & \textbf{0.862} & 0.804 & 0.831 & 0.783 & 0.826 & 0.806 & 0.804 & 0.863 & 0.780 & 0.814 \\
SummerLab-AI         & 0.858          & 0.804 & 0.833 & 0.783 & 0.806 & 0.811 & 0.795 & 0.868 & 0.784 & 0.818 \\
AIA-Noobs            & 0.835          & 0.784 & 0.795 & 0.751 & 0.774 & 0.796 & 0.787 & 0.833 & 0.753 & 0.778 \\
NCC-Next             & 0.831          & 0.761 & 0.809 & 0.768 & 0.775 & 0.786 & 0.755 & 0.833 & 0.735 & 0.787 \\
TSO22                & 0.826          & 0.745 & 0.810 & 0.750 & 0.766 & 0.785 & 0.778 & 0.837 & 0.704 & 0.799 \\
TheOne-Lab           & 0.805          & 0.740 & 0.796 & 0.747 & 0.789 & 0.766 & 0.761 & 0.829 & 0.738 & 0.770 \\
Orsi-Academy         & 0.739          & 0.607 & 0.727 & 0.440 & 0.484 & 0.466 & 0.488 & 0.754 & 0.523 & 0.439 \\
Medical-Mechatronics & 0.506          & 0.378 & 0.476 & 0.356 & 0.270 & 0.361 & 0.247 & 0.445 & 0.344 & 0.291   
\end{tabular}
\label{table:seg_iou}
\end{table*}


\begin{table*}[bth!]
\centering
\caption{Per-video segmentation mNSD~($\uparrow$) results. Scores highlighted in bold show the top-performing method for each test video.}
\begin{tabular}{l|cccccccccc}
Team                & 41    & 42    & 43    & 44    & 45    & 46    & 47    & 48    & 49    & 50    \\
\hline
Uniandes             & 0.887 & \textbf{0.835} & 0.862 & \textbf{0.825} & \textbf{0.895} & \textbf{0.857} & \textbf{0.866} & 0.900 & \textbf{0.836} & \textbf{0.897} \\
HiLab-2022           & \textbf{0.892} & 0.832 & 0.862 & 0.813 & 0.894 & 0.856 & 0.862 & 0.900 & 0.830 & 0.893 \\
SummerLab-AI         & \textbf{0.892} & 0.830 & \textbf{0.863} & 0.811 & 0.881 & 0.856 & 0.855 & \textbf{0.904} & 0.831 & \textbf{0.897} \\
AIA-Noobs            & 0.861 & 0.813 & 0.821 & 0.787 & 0.832 & 0.843 & 0.854 & 0.869 & 0.796 & 0.853 \\
NCC-Next             & 0.859 & 0.786 & 0.835 & 0.802 & 0.844 & 0.835 & 0.813 & 0.869 & 0.776 & 0.873 \\
TSO22                & 0.853 & 0.774 & 0.832 & 0.780 & 0.823 & 0.825 & 0.837 & 0.869 & 0.746 & 0.872 \\
TheOne-Lab           & 0.819 & 0.755 & 0.813 & 0.773 & 0.848 & 0.791 & 0.817 & 0.855 & 0.772 & 0.840 \\
Orsi-Academy         & 0.667 & 0.526 & 0.650 & 0.371 & 0.407 & 0.393 & 0.412 & 0.659 & 0.440 & 0.373 \\
Medical-Mechatronics & 0.507 & 0.380 & 0.481 & 0.357 & 0.274 & 0.363 & 0.253 & 0.454 & 0.349 & 0.306   
\end{tabular}
\label{table:seg_nsd}
\end{table*}

\begin{table*}[bth!]
\centering
\caption{Per video segmentation ranking stability, computed based on eq. \ref{loss:vss}.}
\begin{tabular}{l|cccccccccc|c}
Team                        & 41 & 42 & 43 & 44 & 45 & 46 & 47 & 48 & 49 & 50 & Average~($\uparrow$) \\
\hline
Uniandes                     & 3  & 1  & 1  & 1  & 1  & 1  & 1  & 2  & 1  & 1  & 1.3     \\
HiLab-2022                   & 2  & 2  & 3  & 2  & 2  & 3  & 2  & 3  & 3  & 3  & 2.5     \\
SummerLab-AI                 & 1  & 3  & 2  & 3  & 3  & 2  & 3  & 1  & 2  & 2  & 2.2     \\
AIA-Noobs                    & 4  & 4  & 6  & 5  & 6  & 4  & 4  & 5  & 4  & 6  & 4.8     \\
NCC-Next                     & 5  & 5  & 4  & 4  & 5  & 5  & 7  & 6  & 5  & 5  & 5.1     \\
TSO22                        & 6  & 6  & 5  & 6  & 7  & 6  & 5  & 4  & 7  & 4  & 5.6     \\
TheOne-Lab                   & 7  & 7  & 7  & 7  & 4  & 7  & 6  & 7  & 6  & 7  & 6.5     \\
Orsi-Academy                 & 8  & 8  & 8  & 8  & 8  & 8  & 8  & 8  & 8  & 8  & 8       \\
Medical-Mechatronics         & 9  & 9  & 9  & 9  & 9  & 9  & 9  & 9  & 9  & 9  & 9      
\end{tabular}
\label{table:seg_rank}
\end{table*}

\begin{table}[bth!]
\centering
\caption{Final segmentation results. Scores are calculated following the metrics presented in Sec.~\ref{sec:seg_metrics}. Scores highlighted in bold show the top-performing method for each metric.}\label{table:seg_all}
\begin{tabular}{l|cc|c}
Team                & IoU~($\uparrow$)   & NSD~($\uparrow$)   & Final Score~($\uparrow$) \\
\hline
Uniandes             & \textbf{0.829} & \textbf{0.866} & \textbf{0.847}       \\
HiLab-2022           & 0.817 & 0.863 & 0.840       \\
SummerLab-AI         & 0.816 & 0.862 & 0.839       \\
AIA-Noobs            & 0.789 & 0.833 & 0.811       \\
NCC-Next             & 0.784 & 0.829 & 0.806       \\
TSO22                & 0.780 & 0.821 & 0.800       \\
TheOne-Lab           & 0.774 & 0.808 & 0.791       \\
Orsi-Academy         & 0.567 & 0.490 & 0.527       \\
Medical-Mechatronics & 0.367 & 0.372 & 0.370      
\end{tabular}
\end{table}



The final rank \ref{table:seg_all} shows very similar performance among the top three submissions, all of which are attention-based. The fourth submission was an ensemble of two attention-based networks and one CNN. Teams ranked 5 and 6 submitted attention-based networks and the rest of the submissions were CNN-based. 

Test-time augmentation (TTA) techniques were used in half of the submissions and seemed to effectively increase prediction performance.
AIA-Noobs, 4th place, submitted a CNN-based architecture leveraging TTA method outperforming transformer-based approaches ranked 5th and 6th that didn't implement TTA.
The submission of HiLab in 2nd place, composed by Swin Transformer and UpperNet again leveraging TTA. NCC-Next in the 5th place, used the same architecture as HiLab as part of their submitted network ensemble and achieved lower accuracy. Furthermore, NCC-Next reported that individual models performed worse than their final ensemble. While it is hard to compare the two submissions in this instance, test-time augmentation proved to be a more effective solution compared to a network ensemble at increasing prediction accuracy. 
Furthermore, TheOne-Lab trained their approach from scratch only on SAR-RARP50 and used TTA to achieve a relatively high overall score. 

Every team performed spatial augmentation during training. Teams ranked at positions 2, 5, and 7 performed colour augmentations during training. 

All teams except TheOne-Lab leveraged pre-trained networks before finetuning with SAR-RARP50 data. Uniandes, ranked 1st,  was the only team that used surgical tool datasets in their pretraining phase, which seems to have a major outcome, according to the final ranking.

The top 8 submissions performed excellently in our qualitative evaluation and were generally able to produce accurate instrumentation masks. In the following analysis, we ignore the model submitted by Medical-Mechatronics as it scored significantly lower compared to the other terms. Fig.\ref{fig:res_seg_1} and \ref{fig:res_seg_5} illustrate two of the most interesting and hardest samples in our dataset, according to the IoU scores achieved by the top participant teams.

\def\cwmo{0.38}
\def\cwms{0.7}

\begin{figure*}[bth!]
    \begin{minipage}[c]{\cwmo\textwidth}
        \begin{subfigure}[c]{\textwidth}
            \includegraphics[width=0.9\textwidth]{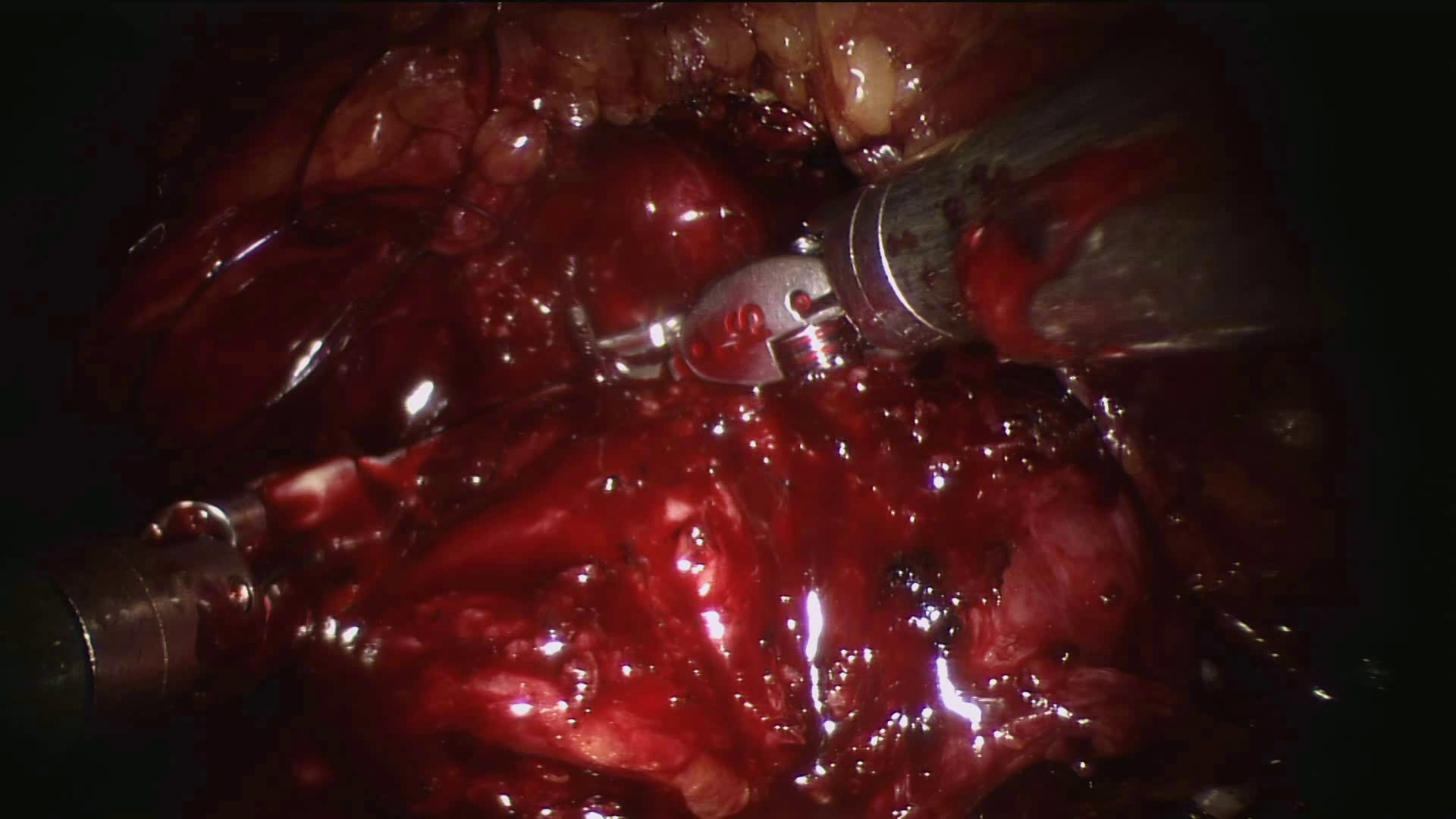}
        \caption{RGB Frame}
        \end{subfigure}
        \begin{subfigure}[c]{\textwidth}
            \includegraphics[width=0.9\textwidth]{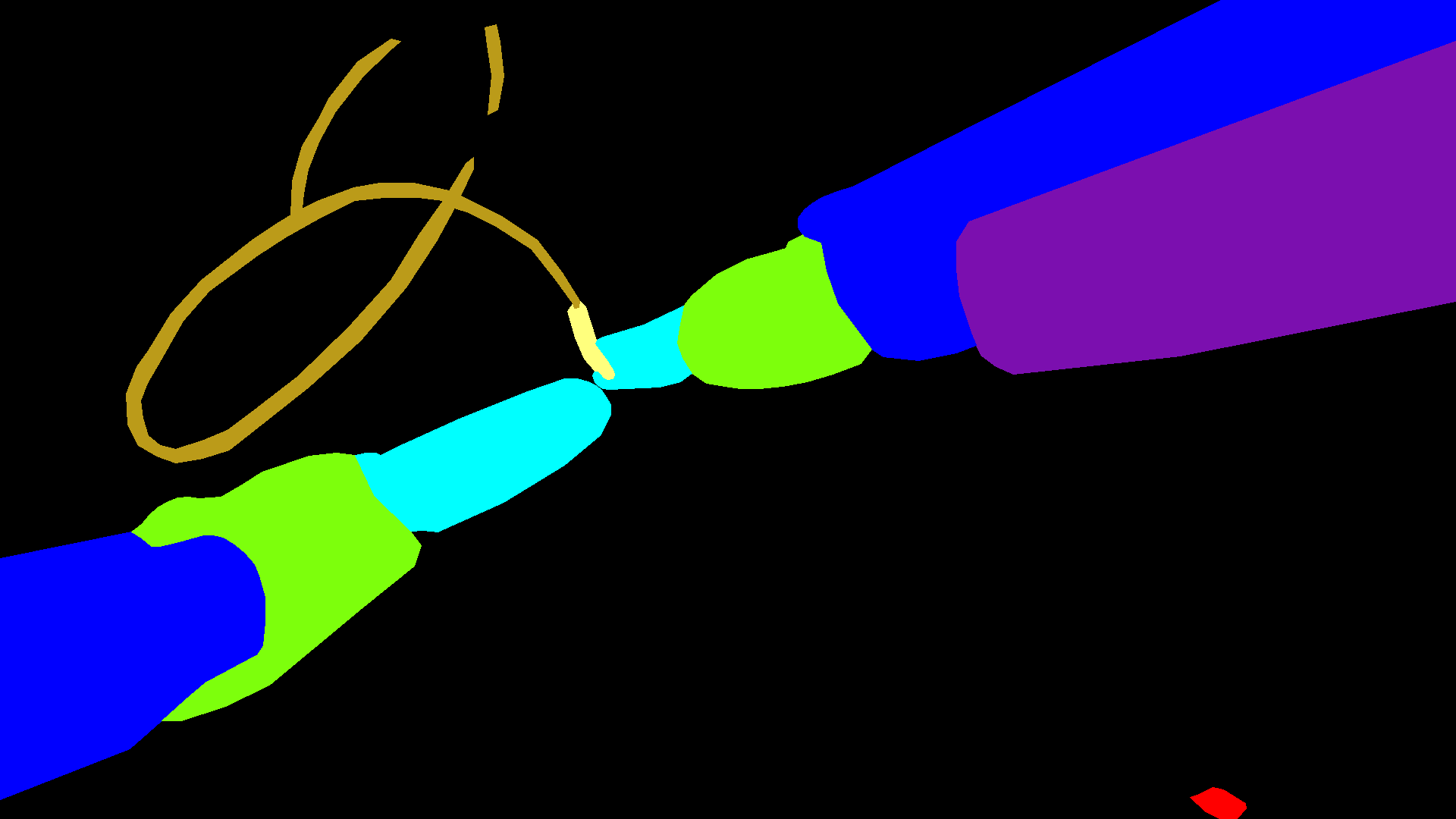}
        \caption{Reference}
        \end{subfigure}
    \end{minipage}%
    \begin{minipage}[c]{\cwms\textwidth}
        \begin{subfigure}[c]{0.3\textwidth}
            \includegraphics[width=\textwidth]{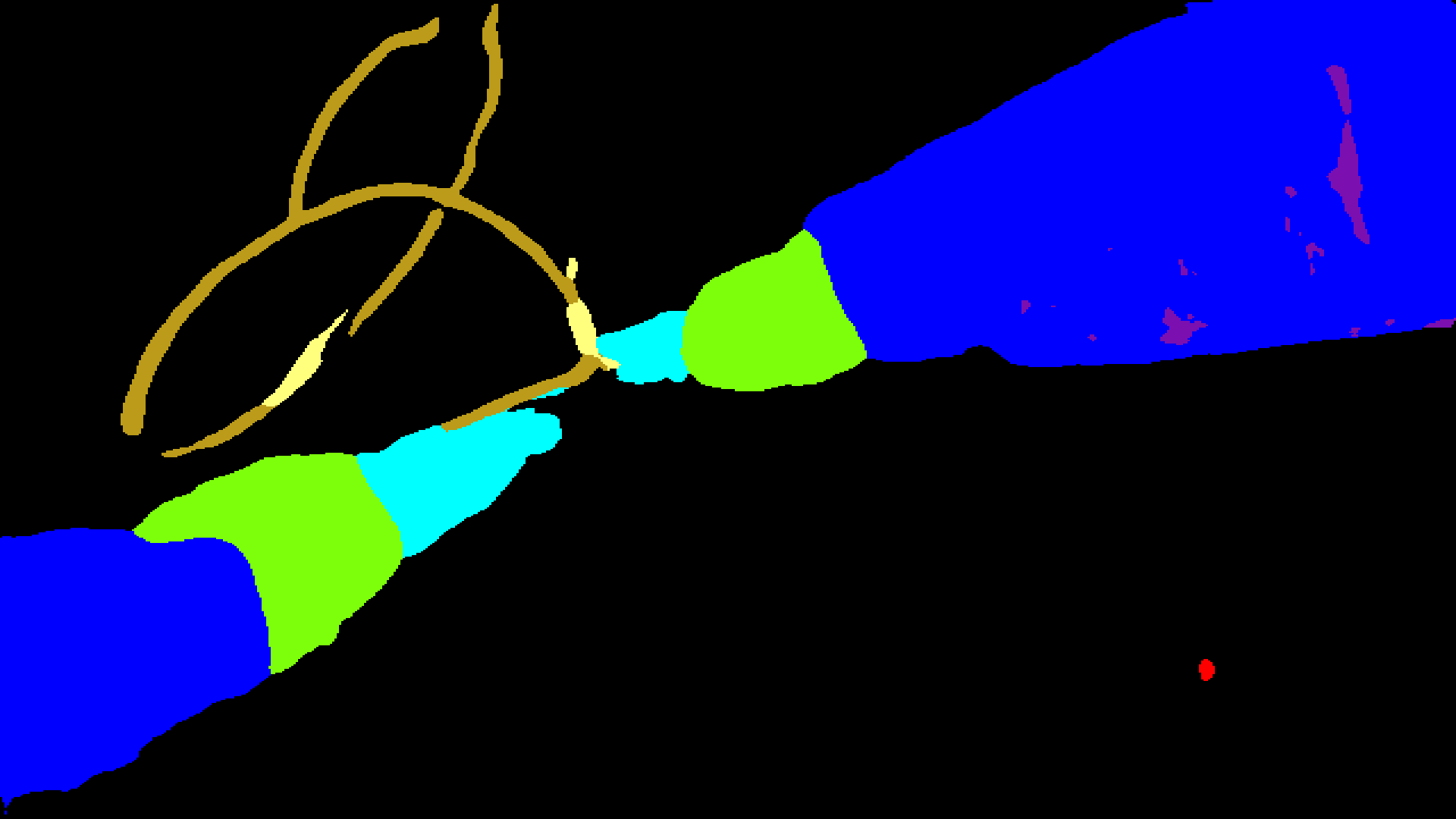}
            \caption{AIA-Noobs}
        \end{subfigure}
        \begin{subfigure}[c]{0.3\textwidth}
            \includegraphics[width=\textwidth]{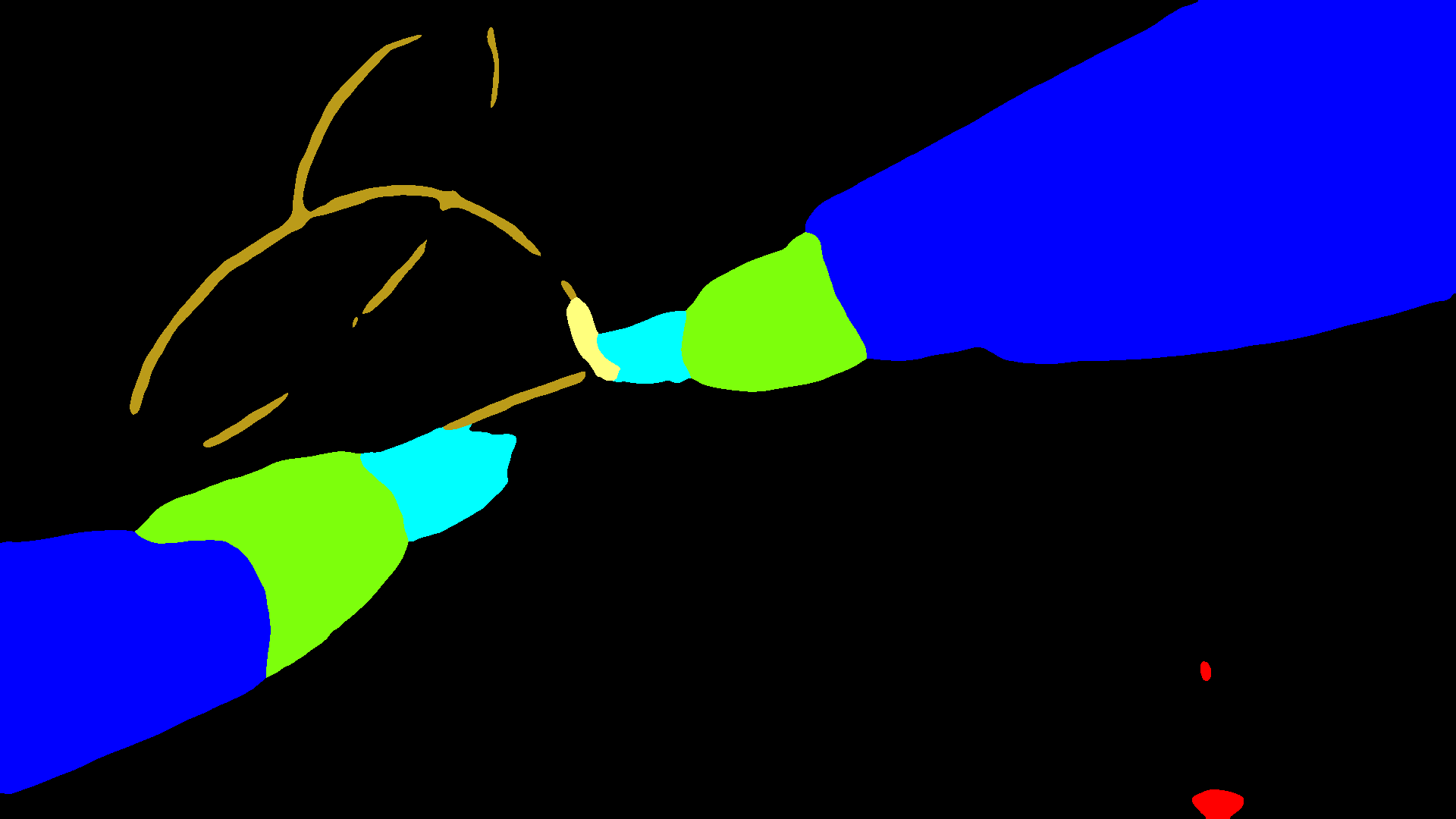}
            \caption{HiLab-2022}
        \end{subfigure}
        \begin{subfigure}[c]{0.3\textwidth}
            \includegraphics[width=\textwidth]{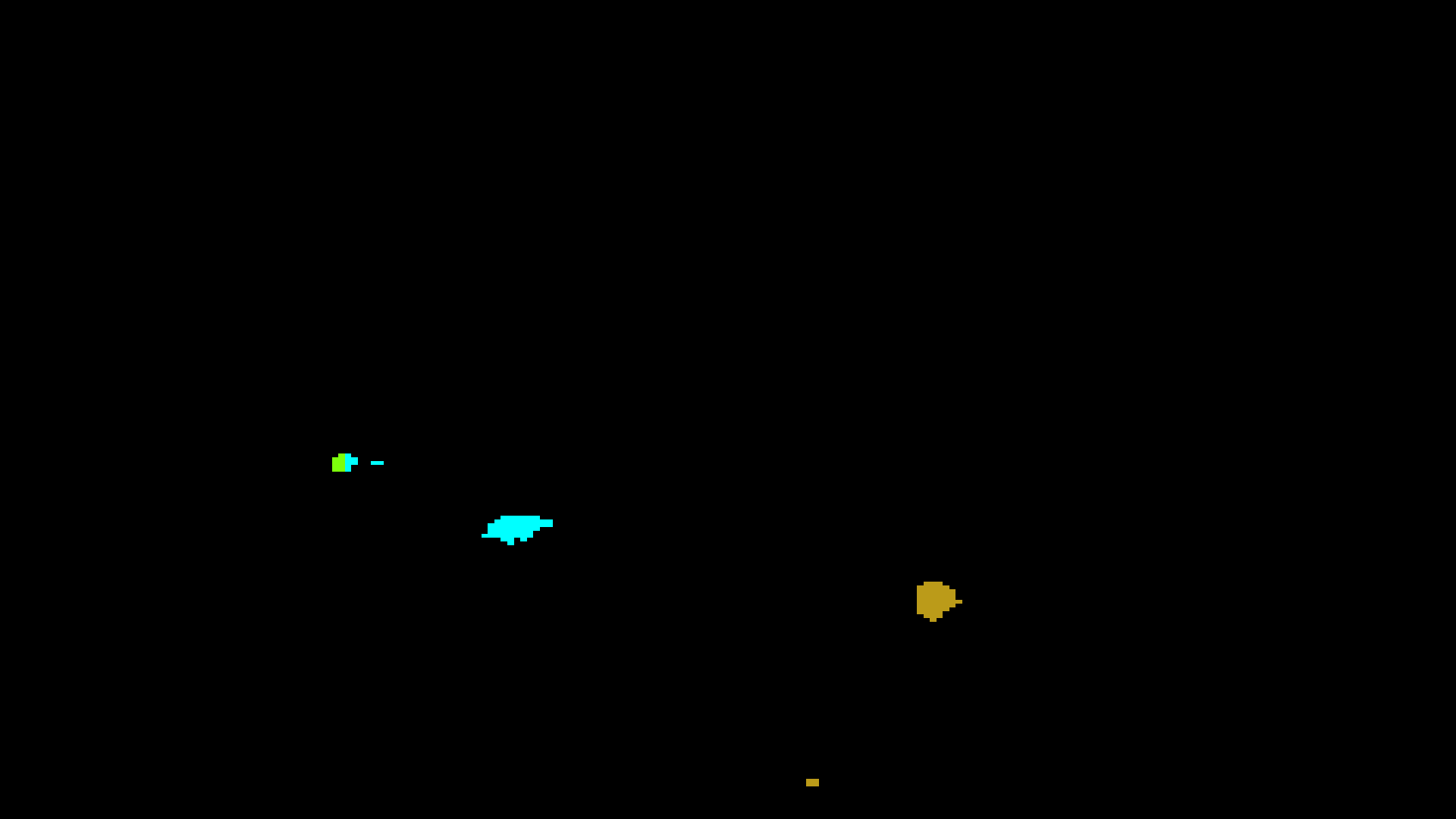}
            \caption{Medical-Mechatronics}
        \end{subfigure}
        
        \begin{subfigure}[c]{0.3\textwidth}
            \includegraphics[width=\textwidth]{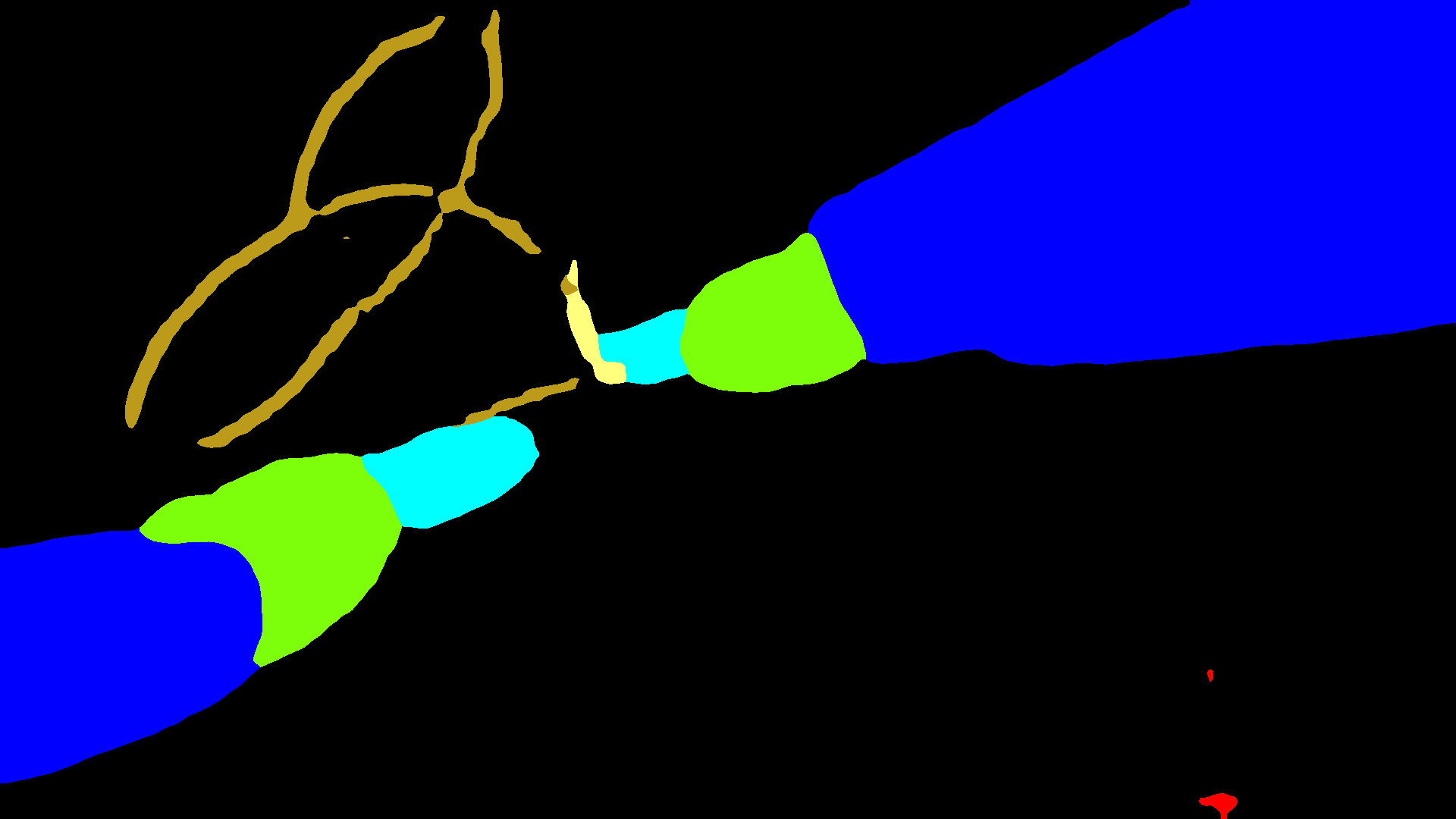}
            \caption{NCC-Next}
        \end{subfigure}
        \begin{subfigure}[c]{0.3\textwidth}
            \includegraphics[width=\textwidth]{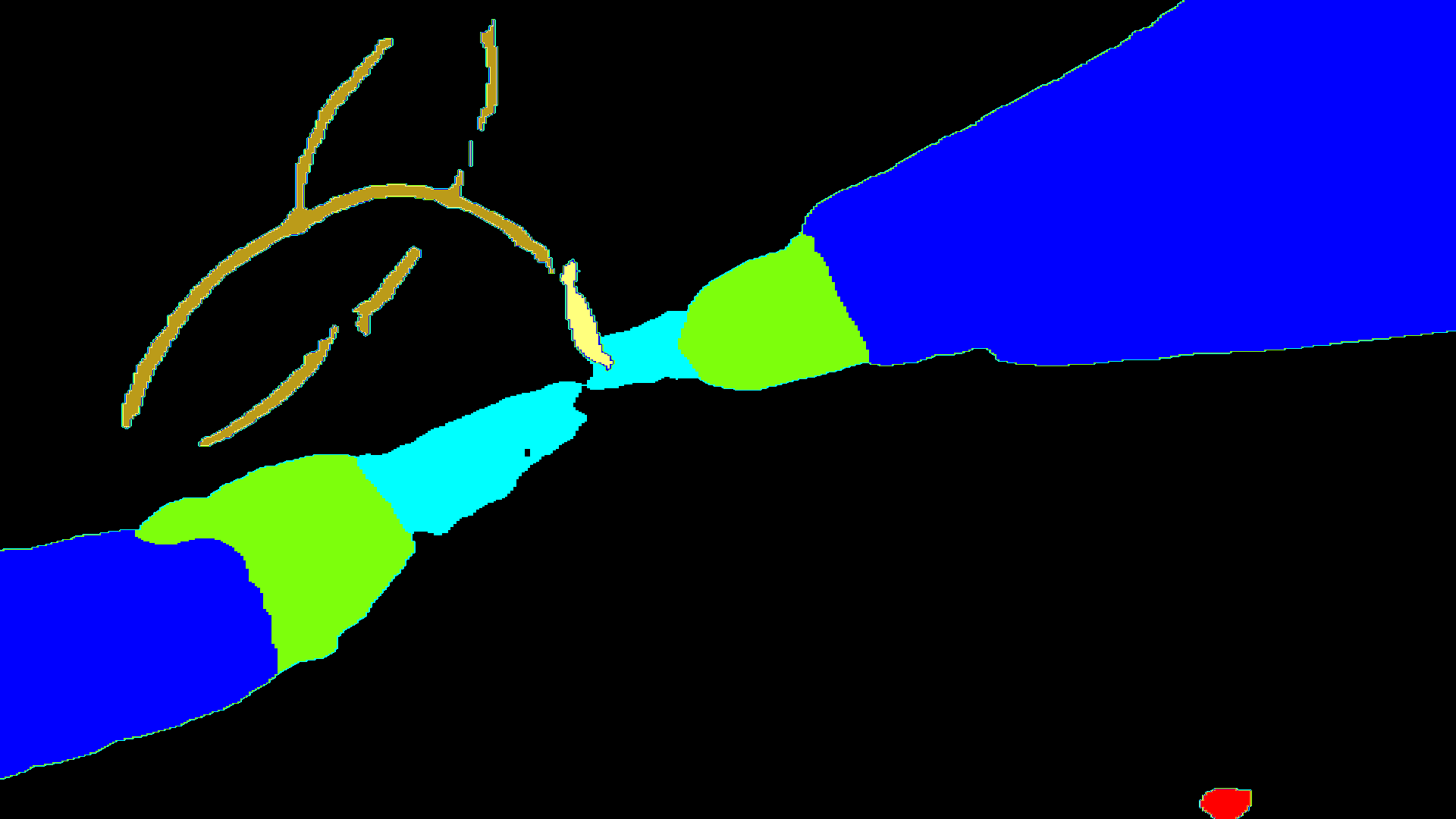}
            \caption{Orsi-Academy}
        \end{subfigure}
        \begin{subfigure}[c]{0.3\textwidth}
            \includegraphics[width=\textwidth]{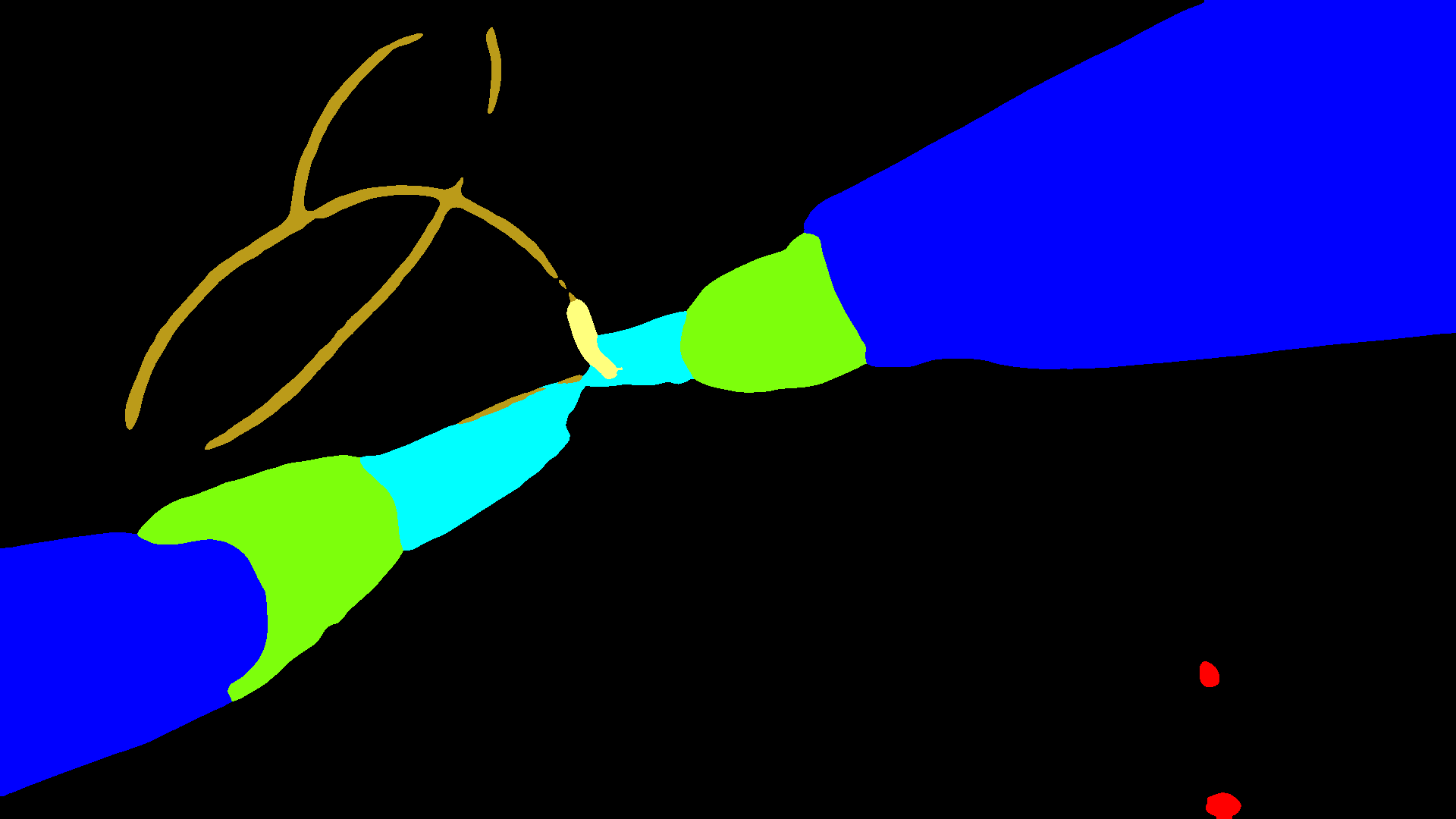}
            \caption{SummerLab-AI}
        \end{subfigure}
        
        \begin{subfigure}[c]{0.3\textwidth}
            \includegraphics[width=\textwidth]{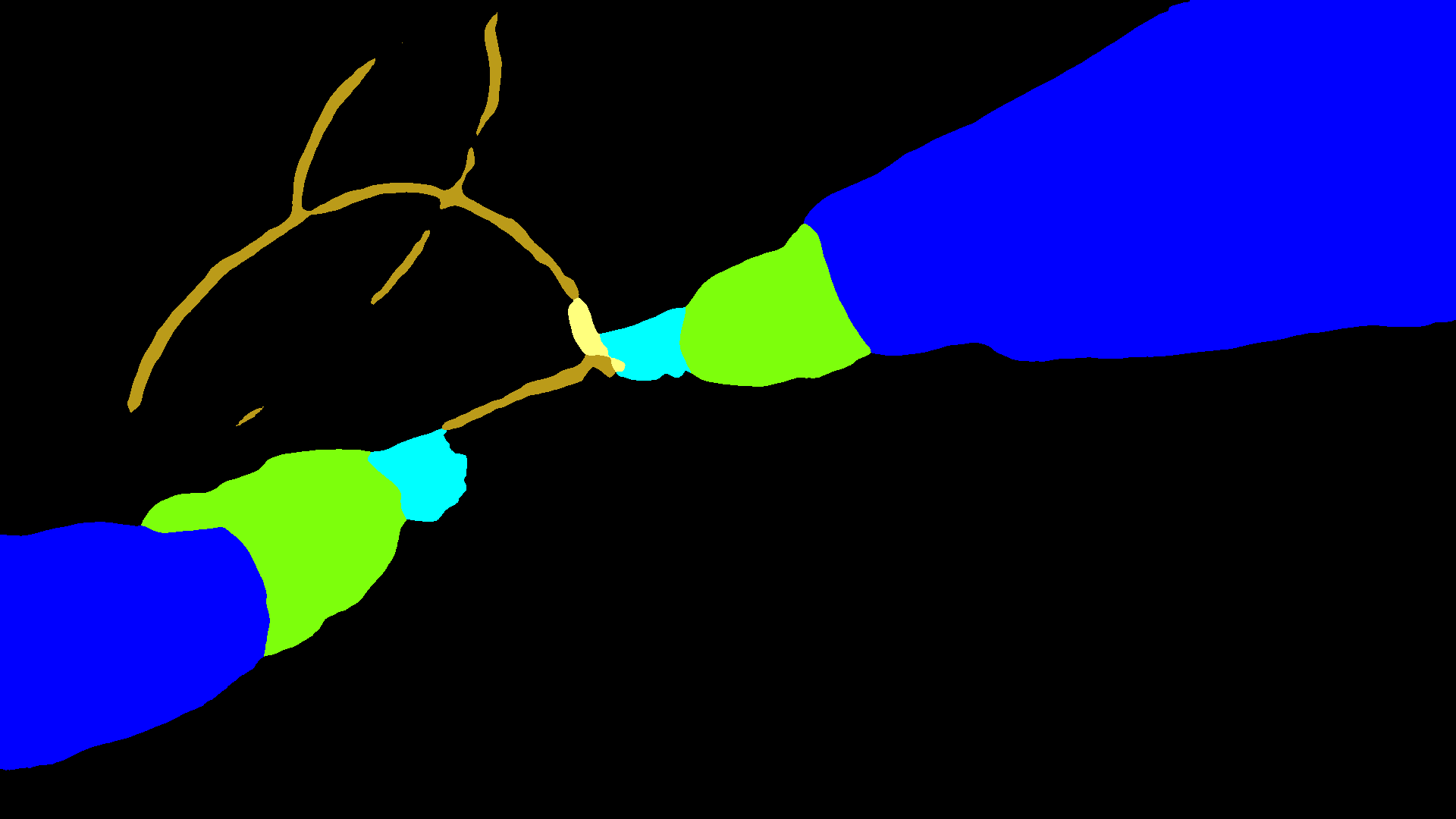}
            \caption{TheOne-lab}

        \end{subfigure}
        \begin{subfigure}[c]{0.3\textwidth}
            \includegraphics[width=\textwidth]{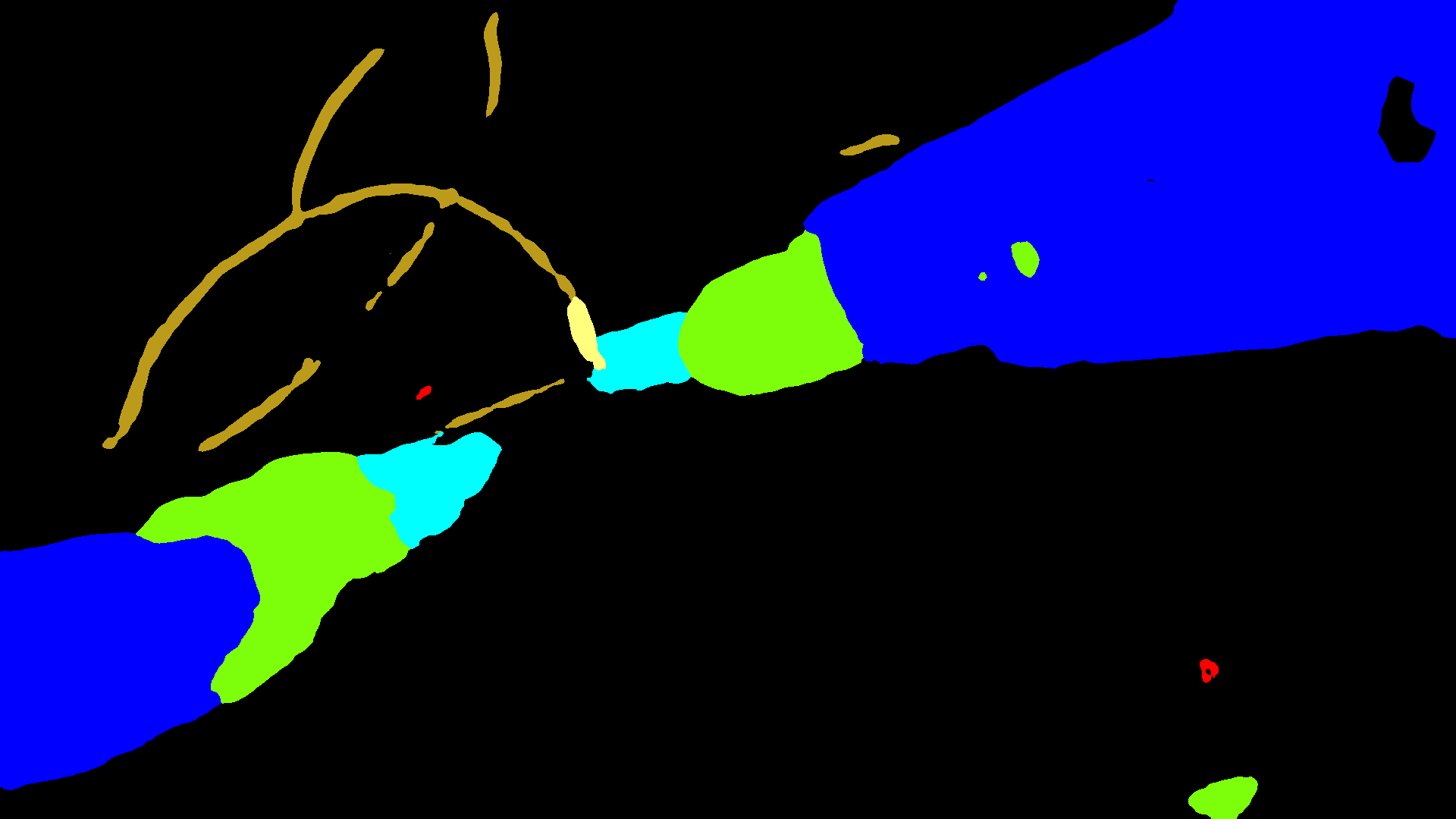}
            \caption{Tso-2022}
        \end{subfigure}
        \begin{subfigure}[c]{0.3\textwidth}
            \includegraphics[width=\textwidth]{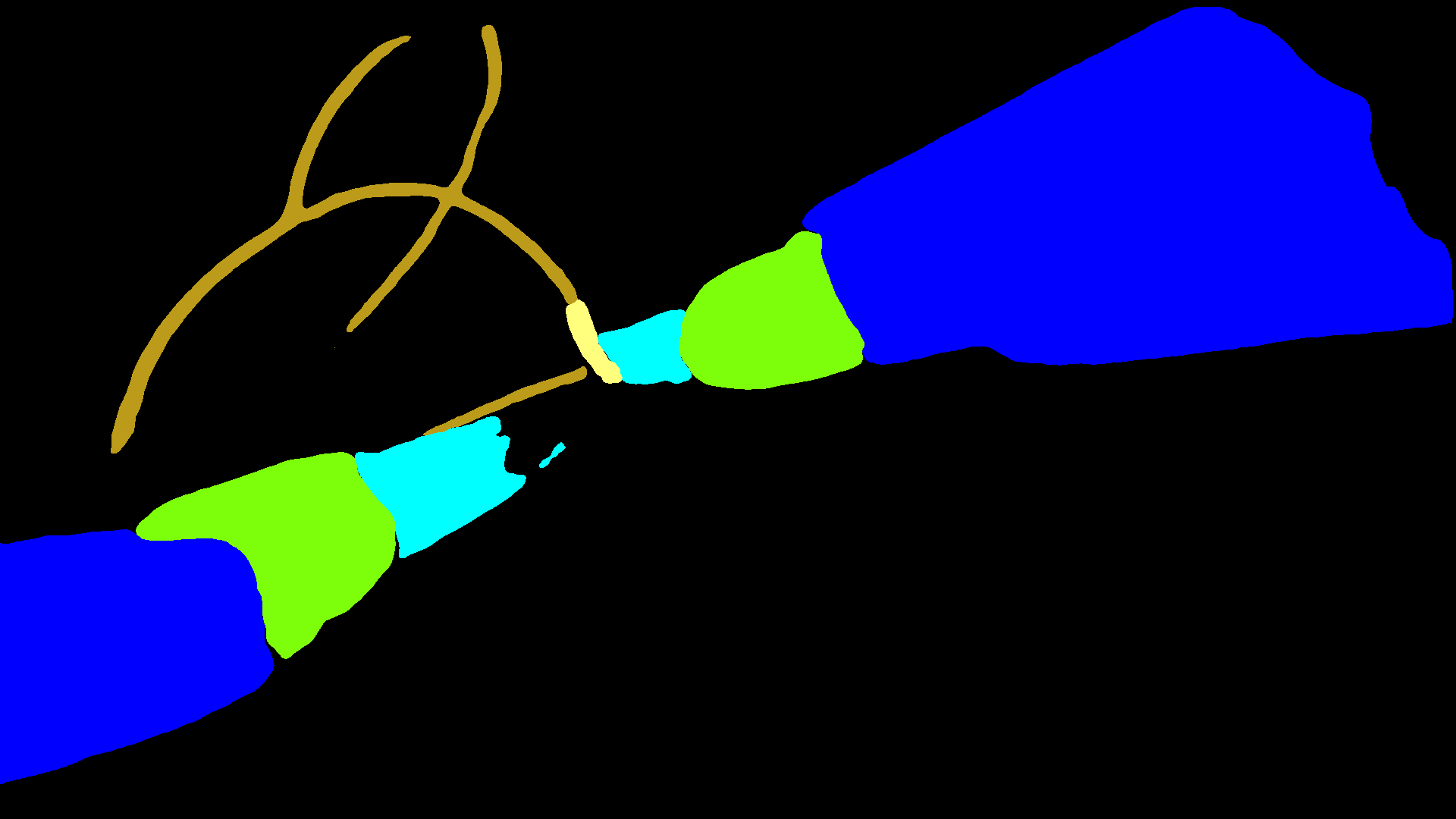}
            \caption{Uniandes}
        \end{subfigure}
  
    \end{minipage}%
    
    \includegraphics[width=0.8\textwidth, center]{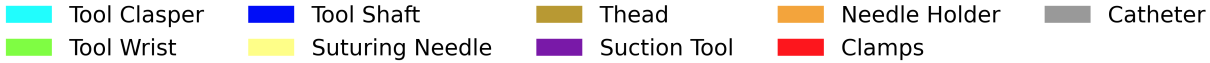}
    
    \caption{Sample predictions from all teams compared to their ground truth. Image from video 42. Most of the proposed models were not able to detect the suction tool.}
    \label{fig:res_seg_1}
\end{figure*}

\begin{figure*}[bth!]
    \begin{minipage}[c]{\cwmo\textwidth}
        \begin{subfigure}[c]{\textwidth}
            \includegraphics[width=0.9\textwidth]{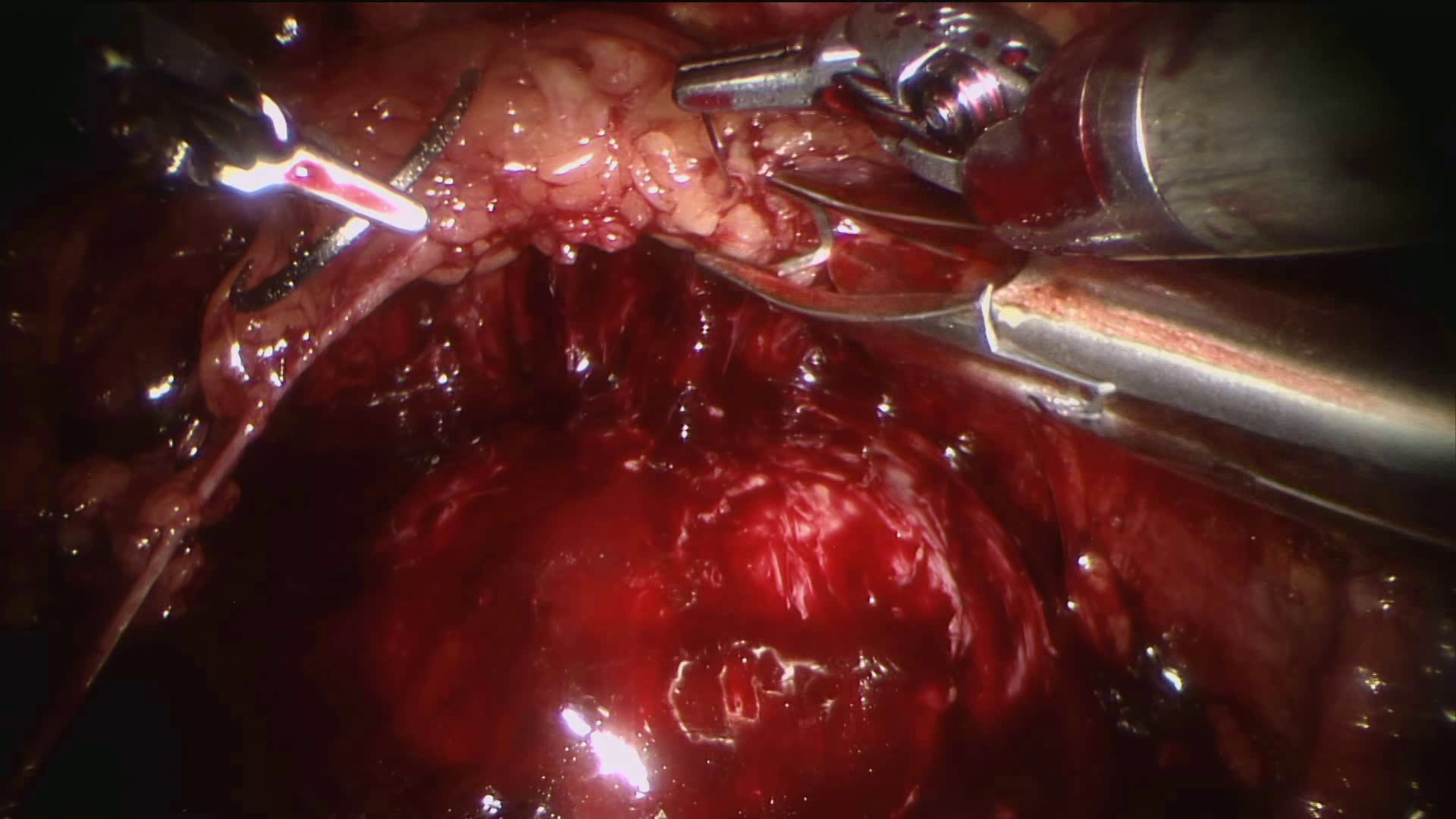}
        \caption{RGB Frame}
        \end{subfigure}
        \begin{subfigure}[c]{\textwidth}
            \includegraphics[width=0.9\textwidth]{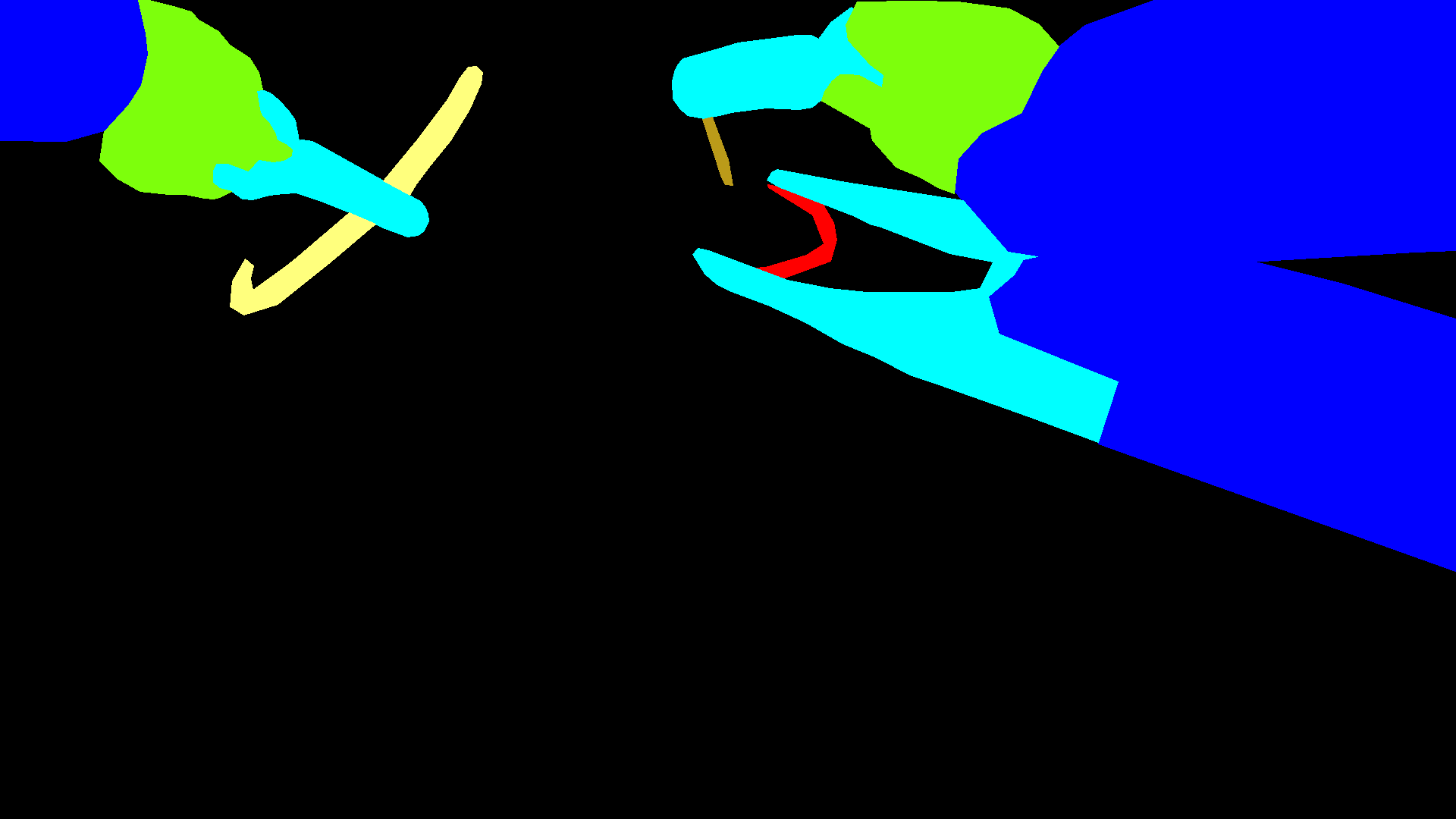}
        \caption{Reference}
        \end{subfigure}
    \end{minipage}%
    \begin{minipage}[c]{\cwms\textwidth}
        \begin{subfigure}[c]{0.3\textwidth}
            \includegraphics[width=\textwidth]{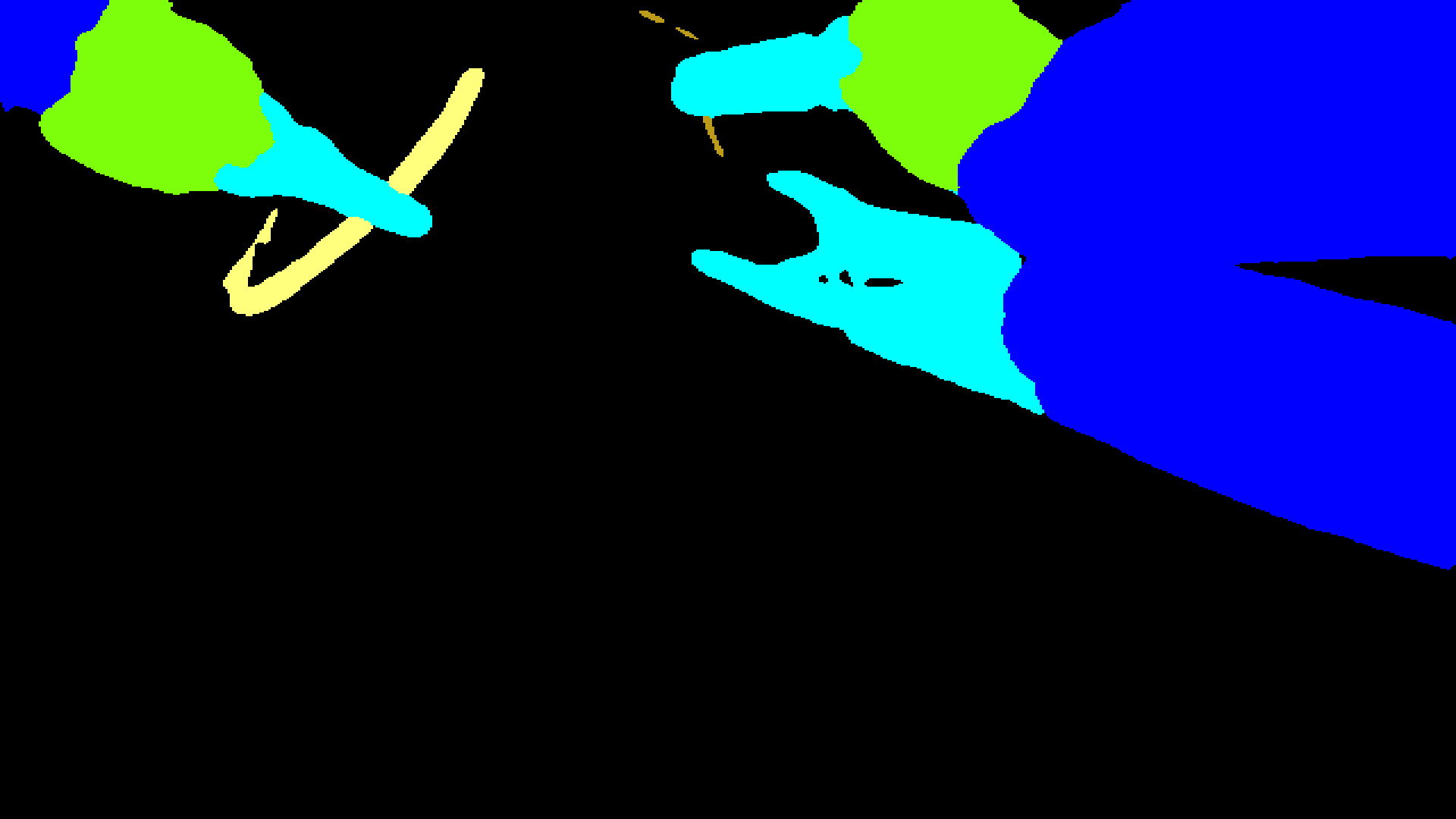}
            \caption{AIA-Noobs}
        \end{subfigure}
        \begin{subfigure}[c]{0.3\textwidth}
            \includegraphics[width=\textwidth]{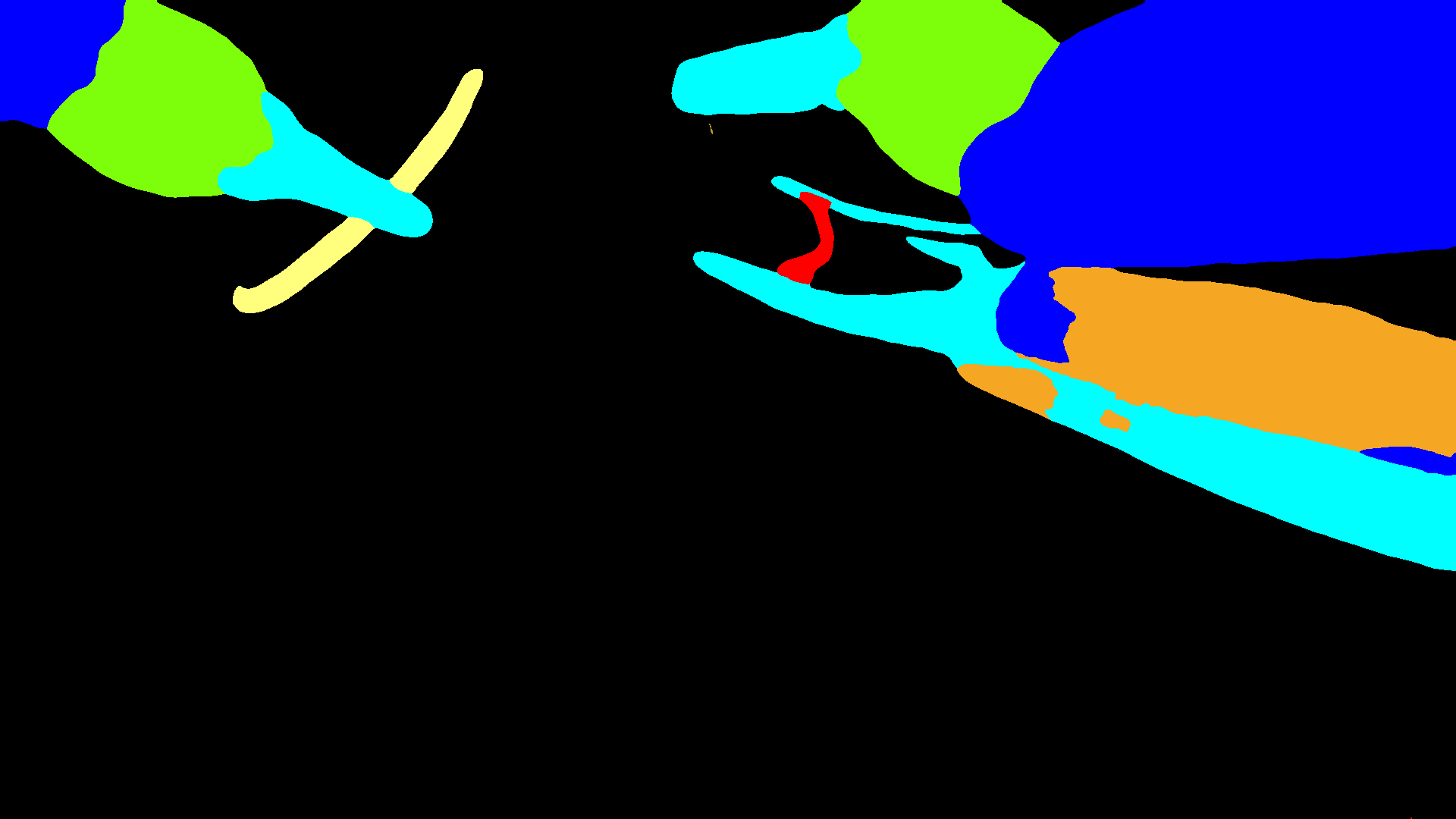}
            \caption{HiLab-2022}
        \end{subfigure}
        \begin{subfigure}[c]{0.3\textwidth}
            \includegraphics[width=\textwidth]{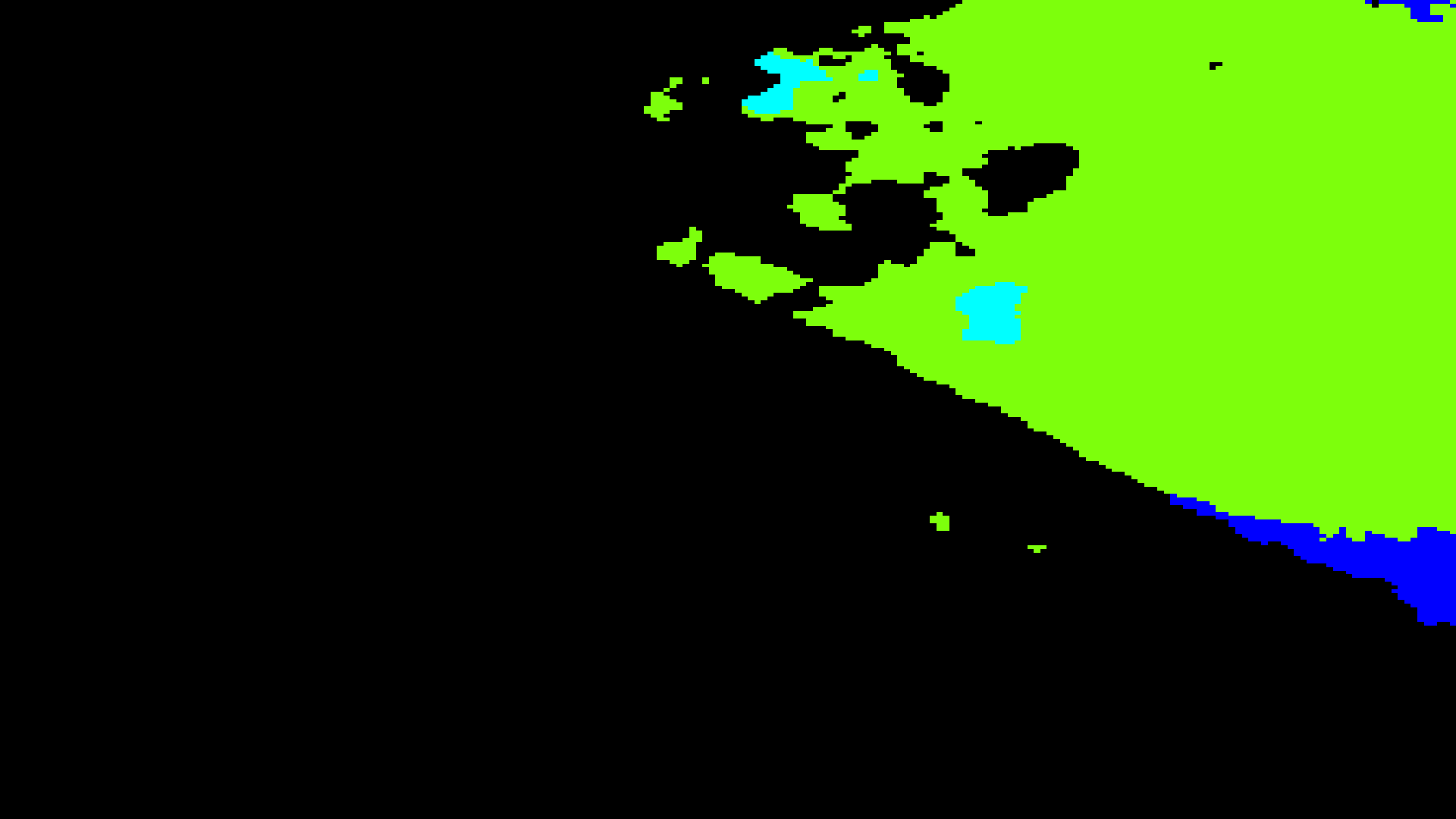}
            \caption{Medical-Mechatronics}
        \end{subfigure}
        
        \begin{subfigure}[c]{0.3\textwidth}
            \includegraphics[width=\textwidth]{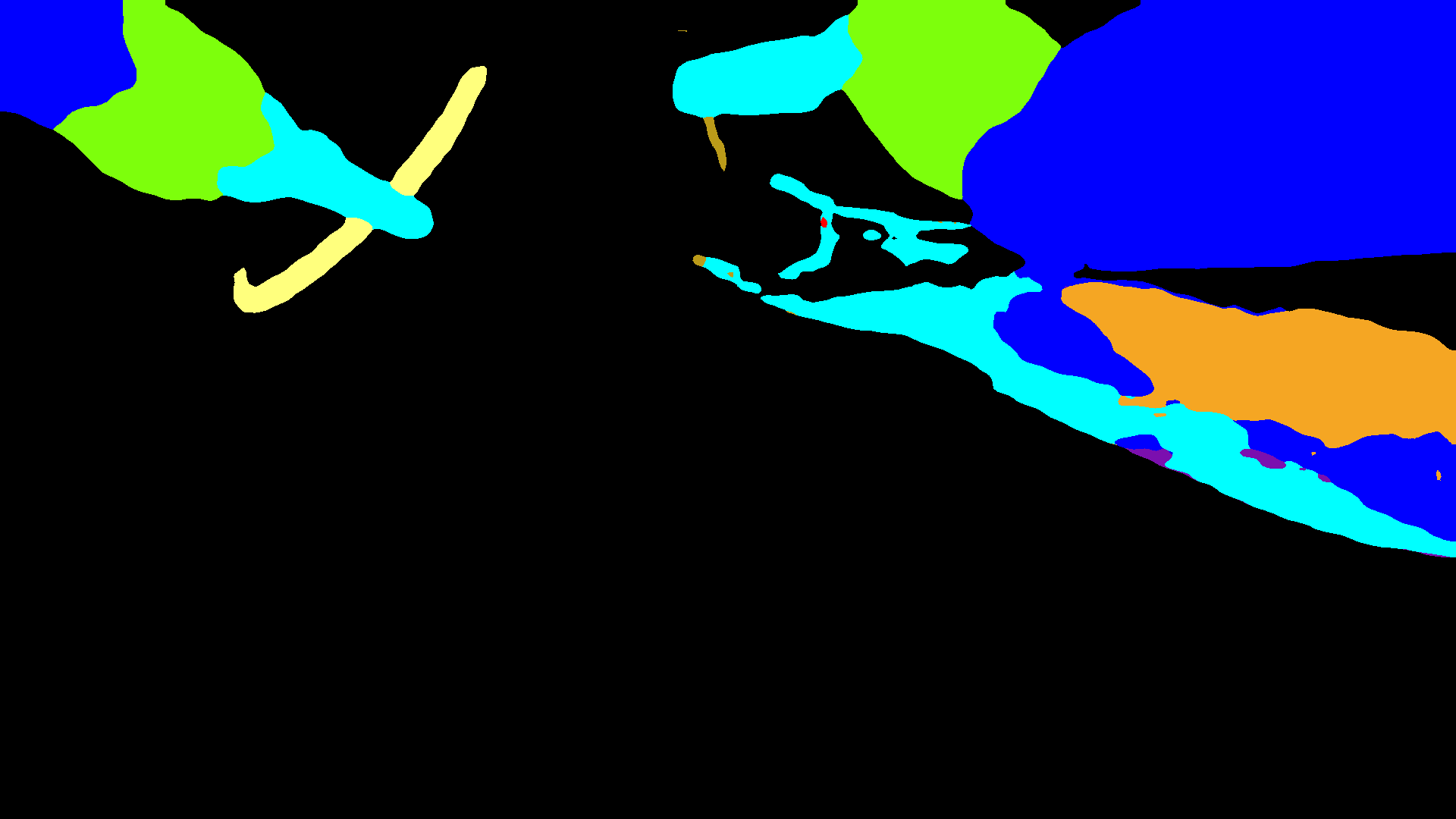}
            \caption{NCC-Next}
        \end{subfigure}
        \begin{subfigure}[c]{0.3\textwidth}
            \includegraphics[width=\textwidth]{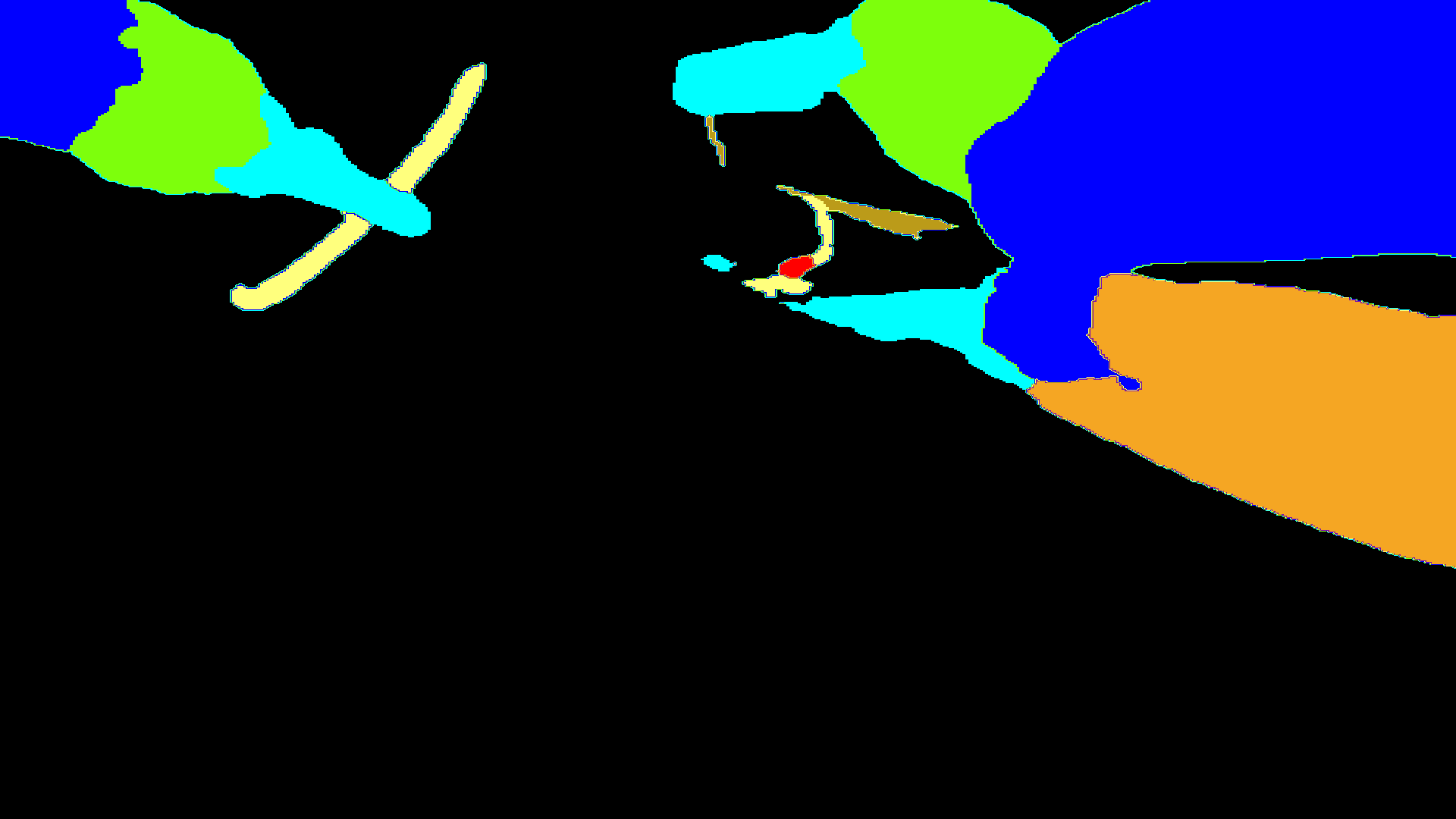}
            \caption{Orsi-Academy}
        \end{subfigure}
        \begin{subfigure}[c]{0.3\textwidth}
            \includegraphics[width=\textwidth]{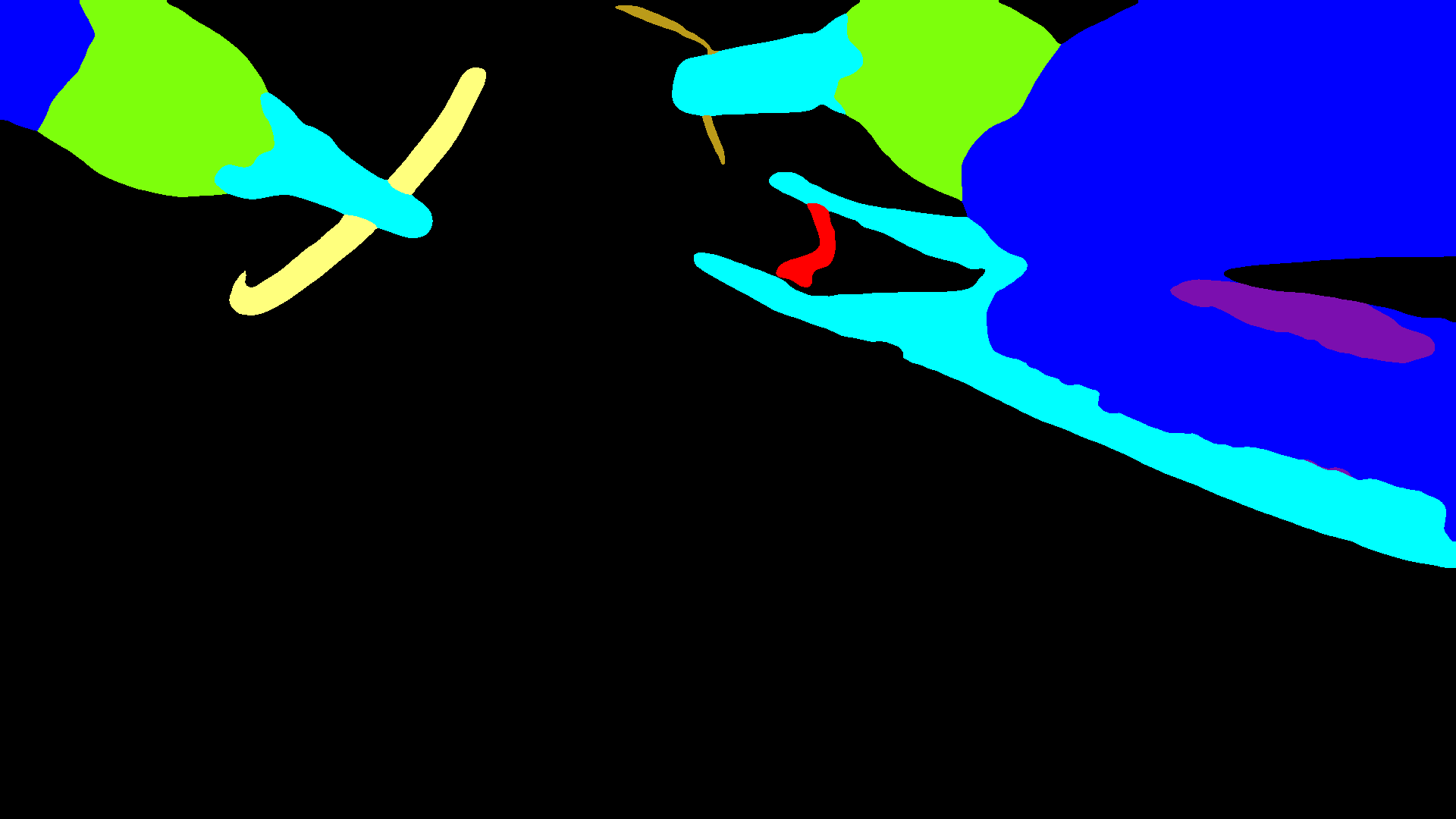}
            \caption{SummerLab-AI}
        \end{subfigure}
        
        \begin{subfigure}[c]{0.3\textwidth}
            \includegraphics[width=\textwidth]{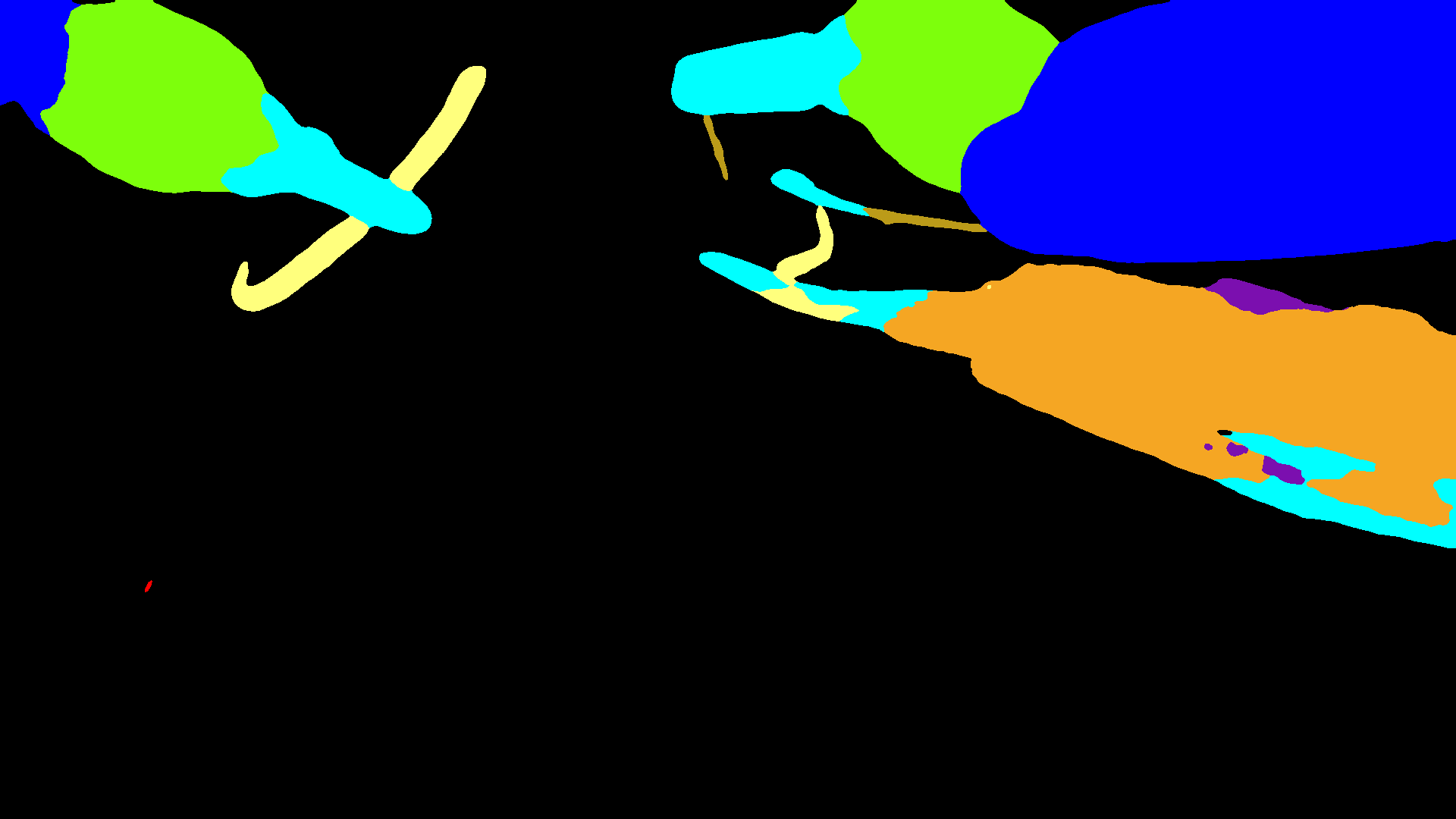}
            \caption{TheOne-lab}

        \end{subfigure}
        \begin{subfigure}[c]{0.3\textwidth}
            \includegraphics[width=\textwidth]{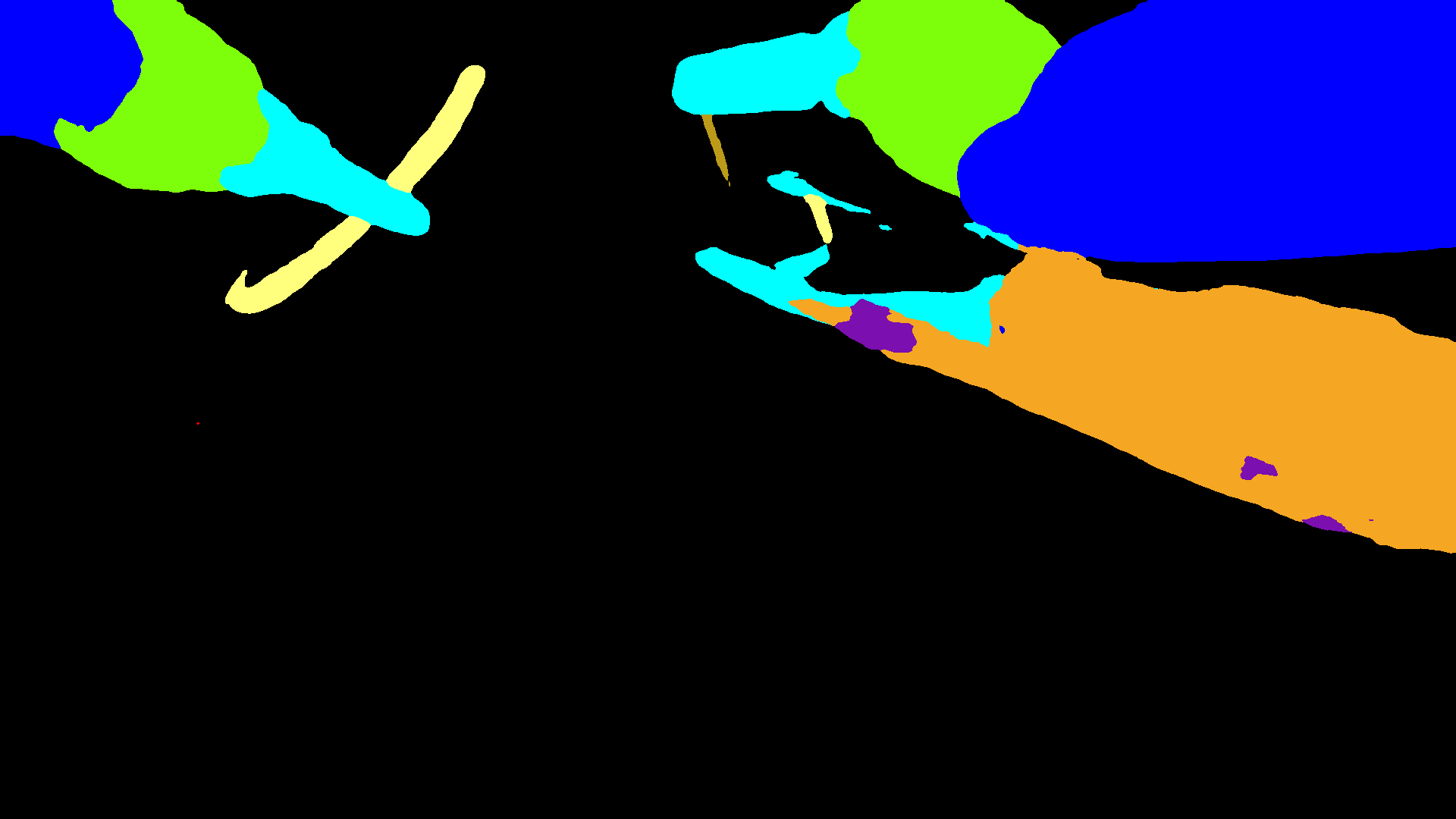}
            \caption{Tso-2022}
        \end{subfigure}
        \begin{subfigure}[c]{0.3\textwidth}
            \includegraphics[width=\textwidth]{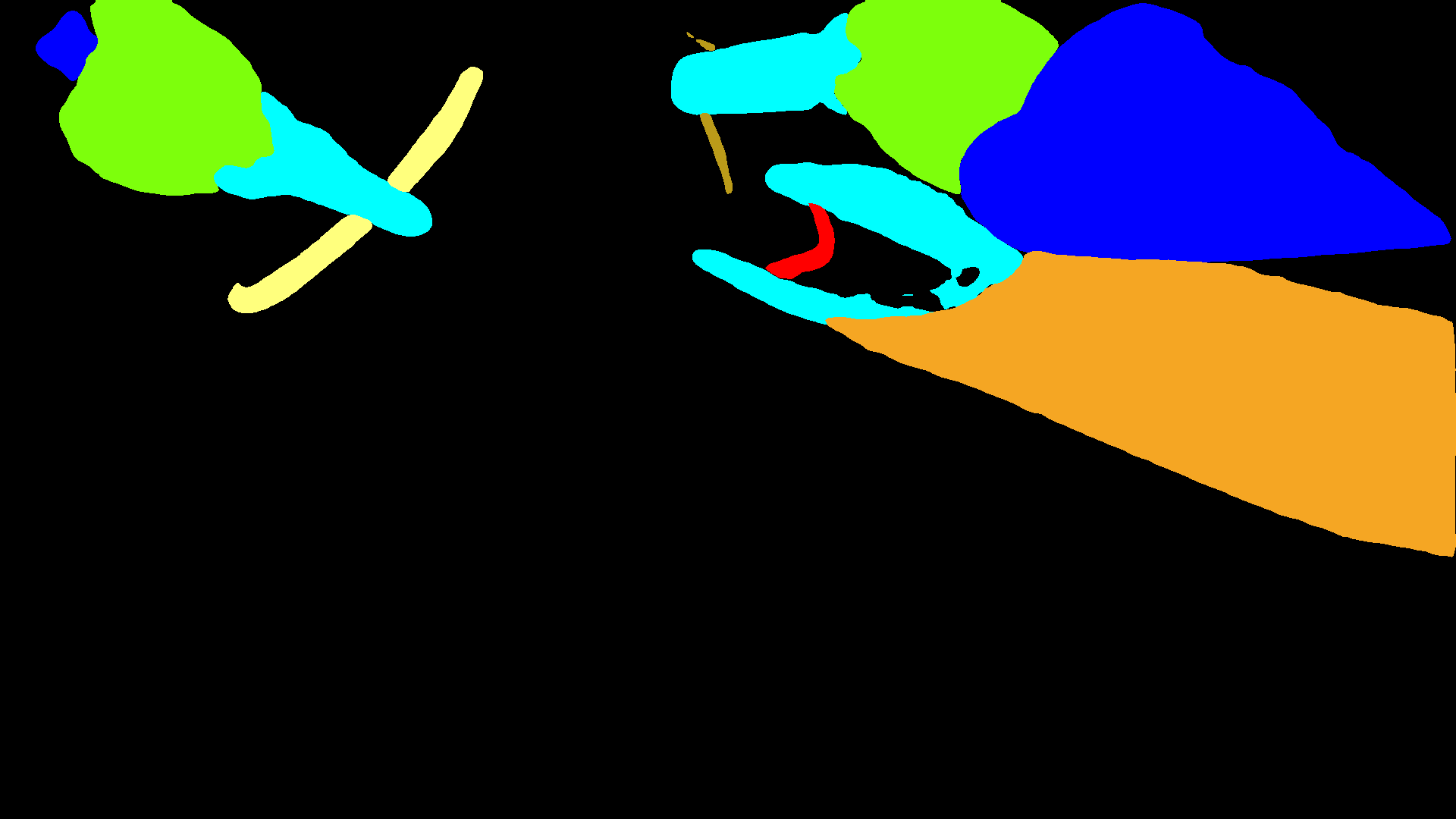}
            \caption{Uniandes}
        \end{subfigure}
  
    \end{minipage}%
    
    \includegraphics[width=0.8\textwidth, center]{figures/results/segmentation/class-colors.png}
    
    \caption{Sample predictions from all teams compared to their ground truth. Image from video 48. Most of the models failed to label the clamp applicator.}
    \label{fig:res_seg_5}
\end{figure*}

The left side of \cref{fig:res_seg_1} includes a tool covered in blood and a thread in a very dark background. All of the methods were able to precisely localize the threads, with the most complete masks being produced by SummerLab-AI and NCC-Next. Interestingly all teams except Orsi-Academy predicted a thread segment over the claspers mask of the left tool which was not observed during the labeling process. The left claspers, which were covered in blood, were partially segmented by most teams but only AIA-Noobs, Orsi-Academy, and SummerLab-AI were able to infer the correct clasper shape. On the right side of the same sample, all teams misclassified the suction tool as a tool shaft. Most teams predicted a mask for the clamp. Finally, all teams except Uniandes propagated the tool shaft mask to the edge of the frame. 

\ref{fig:res_seg_5} is an interesting sample as it includes a clip holder, a tool that appears sporadically across the dataset. All teams, except AIA-Noobs, predicted the wrong class for this tool, which comprises a clasper and a shaft segment. Furthermore, half of the teams segmented the clamp as a needle. 
This behaviour is interesting as it initially points to a bias in certain approaches that mistakenly segment a clamp as a needle. This may occur because the tool manipulating the clamp is inaccurately classified as a needle holder. Second, all approaches predicted a needle holder with a clasper tip. Such a mask is not present in the SAR-RARP50 as the needle holders are segmented as one piece. The semantic classes in the given dataset were chosen to enable accurate part predictions, even for instrumentation that is underrepresented in the dataset. However, in this instance, all methods, except for AIA-Noobs, struggled to generate predictions based on this semantic criterion. 


\subsection{Multi-task sub-challenge}

The results for the multi-task sub-challenge are presented in Table~\ref{table:multi_all}. Action recognition results are presented in Tables ~\ref{table:multi_acc} and ~\ref{table:multi_F10} for Accuracy and F1@10, respectively. Similarly for video segmentation, IoU scores can be found in Table ~\ref{table:multi_iou} and mNSD in Table ~\ref{table:multi_nsd}.

\begin{table*}[bth!]
\centering
\caption{Per Video segmentation mIoU.}
\begin{tabular}{l|ccccccccccc}
Team        & 41    & 42    & 43    & 44    & 45    & 46    & 47    & 48    & 49    & 50    \\
\hline
Uniandes     & \textbf{0.868} & \textbf{0.814} & \textbf{0.853} & \textbf{0.799} & \textbf{0.842} & \textbf{0.815} & \textbf{0.827} & \textbf{0.874} & \textbf{0.799} & \textbf{0.834} \\
AIA-Noobs    & 0.835 & 0.784 & 0.795 & 0.751 & 0.774 & 0.796 & 0.787 & 0.833 & 0.753 & 0.778 \\
SummerLab-AI & 0.764 & 0.708 & 0.757 & 0.721 & 0.729 & 0.703 & 0.676 & 0.771 & 0.674 & 0.689 \\
SK           & 0.696 & 0.655 & 0.735 & 0.634 & 0.680 & 0.706 & 0.655 & 0.761 & 0.612 & 0.695
\end{tabular}
\label{table:multi_iou}
\end{table*}

\begin{table*}[bth!]
\centering
\caption{Per video segmentation mNSD.}
\begin{tabular}{l|ccccccccccc}
Team        & 41    & 42    & 43    & 44    & 45    & 46    & 47    & 48    & 49    & 50    \\
\hline
Uniandes     & \textbf{0.890} & \textbf{0.837} & \textbf{0.871} & \textbf{0.823} & \textbf{0.898} & \textbf{0.852} & \textbf{0.867} & \textbf{0.904} & \textbf{0.836} & \textbf{0.897} \\
AIA-Noobs    & 0.861 & 0.813 & 0.821 & 0.787 & 0.832 & 0.843 & 0.854 & 0.869 & 0.796 & 0.853 \\
SummerLab-AI & 0.761 & 0.711 & 0.765 & 0.733 & 0.776 & 0.727 & 0.737 & 0.785 & 0.695 & 0.754 \\
SK           & 0.713 & 0.669 & 0.752 & 0.660 & 0.719 & 0.739 & 0.702 & 0.779 & 0.638 & 0.760
\end{tabular}

\label{table:multi_nsd}
\end{table*}

\begin{table*}[bth!]
\centering
\caption{Per video action recognition Accuracy.}
\begin{tabular}{l|ccccccccccc}
Team        & 41    & 42    & 43    & 44    & 45    & 46    & 47    & 48    & 49    & 50    \\
\hline
Uniandes     & 0.817 & 0.653 & 0.758 & 0.864 & \textbf{0.829} & \textbf{0.756} & 0.806 & 0.760 & 0.726 & \textbf{0.780} \\
AIA-Noobs    & 0.636 & 0.558 & 0.545 & 0.580 & 0.676 & 0.605 & 0.490 & 0.657 & 0.640 & 0.558 \\
SummerLab-AI & \textbf{0.863} & \textbf{0.717} & \textbf{0.779} & \textbf{0.892} & 0.777 & 0.706 & \textbf{0.809} & \textbf{0.823} & \textbf{0.732} & 0.734 \\
SK           & 0.614 & 0.527 & 0.588 & 0.604 & 0.651 & 0.587 & 0.646 & 0.733 & 0.630 & 0.575
\end{tabular}
\label{table:multi_acc}
\end{table*}

\begin{table*}[bth!]
\centering
\caption{Per video action recognition F1@10.}
\begin{tabular}{l|ccccccccccc}
Team        & 41    & 42    & 43    & 44    & 45    & 46    & 47    & 48    & 49    & 50    \\
\hline
Uniandes     & \textbf{0.879} & \textbf{0.771} & \textbf{0.811} & \textbf{0.923} & \textbf{0.963} & \textbf{0.671} & \textbf{0.790} & \textbf{0.862} & \textbf{0.693} & \textbf{0.867} \\
AIA-Noobs    & 0.714 & 0.544 & 0.676 & 0.769 & 0.724 & 0.549 & 0.491 & 0.723 & 0.584 & 0.578 \\
SummerLab-AI & 0.397 & 0.229 & 0.452 & 0.641 & 0.471 & 0.186 & 0.253 & 0.512 & 0.182 & 0.326 \\
SK           & 0.114 & 0.092 & 0.201 & 0.136 & 0.234 & 0.091 & 0.102 & 0.269 & 0.065 & 0.144
\end{tabular}
\label{table:multi_F10}
\end{table*}

\begin{table*}[bth!]
\centering
\caption{Multi-task ranking stability.}
\begin{tabular}{l|cccccccccc|c}
Team        & 41 & 42 & 43 & 44 & 45 & 46 & 47 & 48 & 49 & 50 & Average \\
\hline
Uniandes     & 1  & 1  & 1  & 1  & 1  & 1  & 1  & 1  & 1  & 1  & 1      \\
AIA-Noobs    & 2  & 2  & 2  & 3  & 2  & 2  & 2  & 2  & 2  & 2  & 2.1    \\
SummerLab-AI & 3  & 3  & 3  & 2  & 3  & 3  & 3  & 3  & 3  & 3  & 2.9    \\
SK           & 4  & 4  & 4  & 4  & 4  & 4  & 4  & 4  & 4  & 4  & 4     
\end{tabular}
\label{table:multi_rank}
\end{table*}

\begin{table*}[bth!]
\centering
\caption{Multi-task final results.}
\begin{tabular}{l|ccc|ccc|c}
Metric       & Accuracy & F1@10 & Action Recognition & IoU & NSD & Segmentation & Final Score \\
\hline
Uniandes     & 0.775   & \textbf{0.823}     & \textbf{0.799}     & \textbf{0.832}     & \textbf{0.868}     & \textbf{0.850}      & \textbf{0.824}       \\
AIA-Noobs    & 0.595   & 0.635     & 0.615     & 0.789     & 0.833     & 0.811      & 0.706       \\
SummerLab-AI & \textbf{0.783}   & 0.365     & 0.534     & 0.719     & 0.744     & 0.732      & 0.625       \\
SK           & 0.615   & 0.145     & 0.299     & 0.683     & 0.713     & 0.698      & 0.456      
\end{tabular}
\label{table:multi_all}
\end{table*}

Overall team Uniandes achieved the best scores for both tasks with \textit{$Score_a$}~=~0.799 and \textit{$Score_s$}~=~0.850, achieving a final score of 0.824. Teams AIA-Noobs and SummerLab-AI achieved second and third places for the multi-task sub-challenge, with a final score of 0.706 and 0.625, respectively. 
The ranking is further confirmed in the stability analysis presented in Table ~\ref{table:multi_rank}.

The top three submissions employed temporal information to make action recognition predictions, while Team-SK's solution relied on single-frame predictions. This lack of temporal context in Team-SK's approach resulted in a substantial performance deficit compared to all other methods, particularly when measuring the F1@10 score.

The participants adopted different strategies to handle the varying sampling rates between segmentation masks and action labels. Uniandes optimized their multitask architecture for each task individually, utilizing all samples from both modalities at their respective sampling rates. AIA-Noobs trained the action recognition component of their network using segmentation priors propagated from 1Hz to 10Hz by replicating the previous segmentation sample for each missing segmentation frame.

Summerlab-AK and Team-SK downsampled the action recognition labels from 10Hz to 1Hz to match the frequency of the segmentation labels and simultaneously trained their multitask models for both tasks. Uniandes and AIA-Noobs, who employed action recognition samples at 10Hz, achieved superior action recognition scores compared to the other two teams.

A direct comparison between single-task and multitask approaches is possible for Uniandes, as they were the only team to combine their single-task architecture with multi-modal data for training. Their multi-task approach achieved a marginally higher segmentation score (0.85) compared to their single-task submission (0.847). In contrast, their multitask model yielded a slightly lower score (0.799) compared to their single-task network (0.804). Overall, their single-task and multi-task approaches exhibited nearly identical performance. AIA-Noobs's multitask submission achieved the same single-task segmentation score because they fed the predictions from their single-task model into their action recognition model, essentially constructing a network cascade with intermediate segmentation predictions. SummerLab-AI opted not to combine their single-task approaches and submitted a network distinct from their single-task submission due to time constraints. Lastly, Team-SK did not submit models trained on single modalities.

The analysis above highlights the challenges associated with employing multitask approaches and multi-modal data for optimization. Team Uniandes's multitask submission did not demonstrate any significant advantage over their single-task submission. Therefore, based on the submitted solutions, it remains inconclusive whether the multi-modal nature of SAR-RARP50 can be leveraged to enhance model accuracy in a multitask learning scenario.

\section{Conclusions}

The Endovis2023 SAR-RARP50 challenge introduced a video dataset of real RARP procedures, along with reference surgical gesture and semantic instrument annotations. The challenge aimed to promote advancements in surgical action recognition and surgical tool segmentation, while also exploring the potential benefits of jointly addressing these tasks.

Twelve teams participated in the challenge, submitting seven action recognition, nice instrument segmentation, and four multi-task solutions. Learning-based models, combining attention and convolution, were prevalent. Top-performing solutions incorporated attention mechanisms, with two-stage action recognition approaches proving highly accurate. Post-processing techniques, like prediction filtering, significantly enhanced action recognition performance. Notably, a drop in accuracy occurred for videos from less experienced surgeons, indicating deviations from standardized workflows.

Instrumentation segmentation methods achieved high accuracy, successfully predicting tool and thread masks, even in blood-obscured scenarios. Test-time augmentation was shown to improve segmentation, but predicting smaller objects and underrepresented classes posed challenges, revealing biases in some methods.


The direct comparisons between multi-task submissions and single-task solutions proved difficult due to factors such as teams not extending their single-task architectures to multi-task or not fully utilizing the multi-modal nature of the dataset in a joint optimization scheme. Consequently, based on the received submissions, it remains unclear whether multi-task approaches offer definitive advantages over single-task models.

To enhance the robustness of such systems, future research should explore leveraging existing datasets and additional modalities to empower models to make predictions in previously unobserved types of operations.


\section{Acknowledgements}
This work was supported in part, by the Wellcome/EPSRC Centre for Interventional and Surgical Sciences (WEISS) [203145/Z/16/Z] and the Royal Academy of Engineering under the Chair in Emerging Technologies programme. We thank Intuitive Surgical who generously sponsored the majority of the surgical segmentation annotation process and provided prize awards for the winners of the the SAR-RARP50 challenge.

\bibliographystyle{model2-names.bst}\biboptions{authoryear}
\bibliography{references}

\end{document}